\documentclass[10pt,twocolumn]{article}

\usepackage[T1]{fontenc}
\usepackage{times}
\usepackage{latexsym}
\usepackage{microtype}

\usepackage{amsmath}
\usepackage{amssymb}

\usepackage{graphicx}
\usepackage{booktabs}
\usepackage{multirow}
\usepackage{xcolor}
\usepackage{subcaption}
\usepackage{float}
\usepackage{stfloats}  

\usepackage[breaklinks,pdftitle={AMEL: Accumulated Message Effects on LLM Judgments},pdfauthor={Temkit Sid-Ali},pdfsubject={LLM evaluation bias},pdfkeywords={LLM-as-a-judge, conversation history, bias, evaluation}]{hyperref}
\usepackage[numbers]{natbib}
\usepackage{url}

\usepackage[letterpaper,margin=0.75in,columnsep=0.35in]{geometry}

\newcommand{\cohend}{Cohen's $d$}
\newcommand{\xspace}{}

\newcommand{\eg}{e.g.,\xspace}

\title{AMEL: Accumulated Message Effects on LLM Judgments}

\author{
  Temkit Sid-Ali \\
  chut.app \\
  \texttt{sid@chut.app}
}

\begin{document}
\maketitle

\begin{abstract}
LLMs are now used as judges. They review code, moderate user posts, mark exam answers. Production setups often run many items through one conversation to save tokens, and that is the setup I worry about here. If the model has just said ``no'' ten times, does the eleventh item still get a fair hearing? I test this directly. The effect, when it appears, is what I call the \textit{accumulated message effect on LLM judgments}---AMEL.

The setup is the same in each of three binary-judgment domains (code review, content moderation, nutrition): present each test item in a clean conversation, or after $N$ prior turns whose answers are mostly ``yes'' or mostly ``no,'' and ask whether the verdict shifts. Twelve models across five providers, 84,088 API calls after deduplication.

The shift is small but consistent. Overall $d = -0.17$ with item-clustered 95\% CI $[-0.21, -0.13]$; the underlying mean bias score is $\bar{BS} = -0.054$, i.e., conversation history moves $P(\text{no})$ by roughly five percentage points on average. The effect is concentrated where it bites: items the model is itself uncertain about at baseline absorb roughly twice as much ($d = -0.36$ for high-entropy items vs.\ $d = -0.15$ for items it already calls deterministically). It does not grow with context length: 5 turns and 50 produce the same shift (Spearman $|r| < 0.01$), though I cannot resolve where the saturation actually occurs below $N = 5$. Negative history pulls harder than positive (paired per-item ratio $1.52\times$; the marginal-means ratio of $\sim 2.1\times$ mixes item composition and should not be taken as the headline number). Bigger models help, but only a little. The Nano $\to$ GPT-5.2 step ($-0.34 \to -0.17$) is the credible scaling signal in the panel; the within-Anthropic ladder ($-0.22 \to -0.18 \to -0.17$) is small enough ($\Delta d = 0.05$ from Haiku to Opus) to sit inside the per-model CI width and is at most suggestive.

Three follow-ups try to narrow what is going on. The first-token probability distribution shifts continuously, not at a threshold. The negativity asymmetry has both token-level and semantic flavours, though with $n = 21$ items per cell I cannot tell which dominates per model. Position does not matter: five biased turns anywhere in a 50-turn window produce the same shift. The simplest fix for an evaluation pipeline is a fresh context per item. If you must batch, interleave expected-yes and expected-no items and brace for a residual negativity drift.
\end{abstract}

\section{Introduction}
\label{sec:intro}

Language models grade things now. Code. Comments. Essays. Creative writing \cite{zheng2024judging}. In production, the cheap option---and so usually the chosen one---is to feed many items through one conversation rather than open a fresh session for each.

That setup creates an obvious worry. A code reviewer has just rejected ten pull requests in a row; what happens to submission eleven? Does the accumulated history of ``no'' answers quietly move the goalposts? I tested it.

The design is simple. Each item is presented either alone (baseline) or after $N$ prior turns whose answers are mostly ``yes'' or mostly ``no.'' I then check whether the verdict shifts. I call the shift, when it appears, the \textit{accumulated message effect} (AMEL).

A few clarifications up front. AMEL is not sycophancy \cite{sharma2024understanding}: no user opinion is expressed in the prompts; the model conforms to \textit{its own} previous answers. It is not the majority-label bias of few-shot classification \cite{zhao2021calibrate} either---the prior turns here are full question-answer exchanges on varied topics within the domain, not labelled demonstrations. And it is not anchoring in the \citet{jones2024anchoring} sense: no number is supplied; what shifts the model is the aggregate \textit{tone} of the conversation rather than an explicit reference value.

Main findings:
\begin{enumerate}
    \item Across the 12 tested models (drawn from 5 providers, with 1--4 models per provider), conversation history pulls responses toward its prevailing polarity ($d = -0.17$, $p < 10^{-53}$); 10 of 12 models show a significant effect after Bonferroni correction under the conservative first-occurrence Qwen3 30B deduplication. Under last-occurrence or random-selection dedup the count is 11 of 12 (Qwen3 30B significant contrarian, $d = +0.17$ to $+0.22$). The choice is load-bearing for this single row; the other 11 model classifications are stable across dedup schemes (Appendix~\ref{app:dedup}). The result holds in every provider sub-panel, but at 1--4 models per provider this is a coverage observation, not a provider-level claim.
    \item Susceptibility tracks the model's own baseline uncertainty. Items where the model is genuinely uncertain at baseline (nonzero binary entropy of $P(\text{yes}|\text{baseline})$) absorb $\approx 2\times$ more bias than items where the baseline is deterministic ($d = -0.36$ vs.\ $d = -0.15$). The author-coded ``ambiguous'' label predicts bias size only because some author-ambiguous items are also empirically uncertain; among items the model already finds clear-cut, author labels add little (Section~\ref{sec:empirical_entropy}).
    \item Negative history biases more than positive history: paired per-item $|BS|$ ratio $1.52\times$ ($t = 13.03$, $p < 10^{-36}$, $n = 2{,}733$). Marginal means give $\approx 2.1\times$ but mix item composition across cells; the paired test is the cleaner statistic.
    \item Bias saturates early. Five turns of skewed history produce the same shift as fifty (Spearman $|r| < 0.01$, $p = 0.92$; OLS slope $-4 \times 10^{-5}$ BS/turn, $p = 0.84$). The model recognizes the pattern fast and additional examples do not strengthen it.
    \item Larger models within provider families show less bias, but none are immune.
    \item The effect is detected in all three tested domains, all significant after Bonferroni correction: code review ($d = -0.25$) $>$ meals ($d = -0.13$) $\approx$ content moderation ($d = -0.12$). For content moderation the parametric test is significant but the item-clustered bootstrap CI just touches zero, so the per-domain content-moderation magnitude is the most stimulus-sensitive of the three.
    \item Three characterization experiments (Section~\ref{sec:mechanism}) narrow the plausible mechanisms but do not isolate a single cause: the probability distribution shifts continuously rather than crossing a threshold, the negativity asymmetry has both token-level and semantic components (per-model attribution is exploratory at $n = 21$ items), and position within the conversation does not matter.
\end{enumerate}

\begin{figure*}[t]
    \centering
    \includegraphics[width=\textwidth]{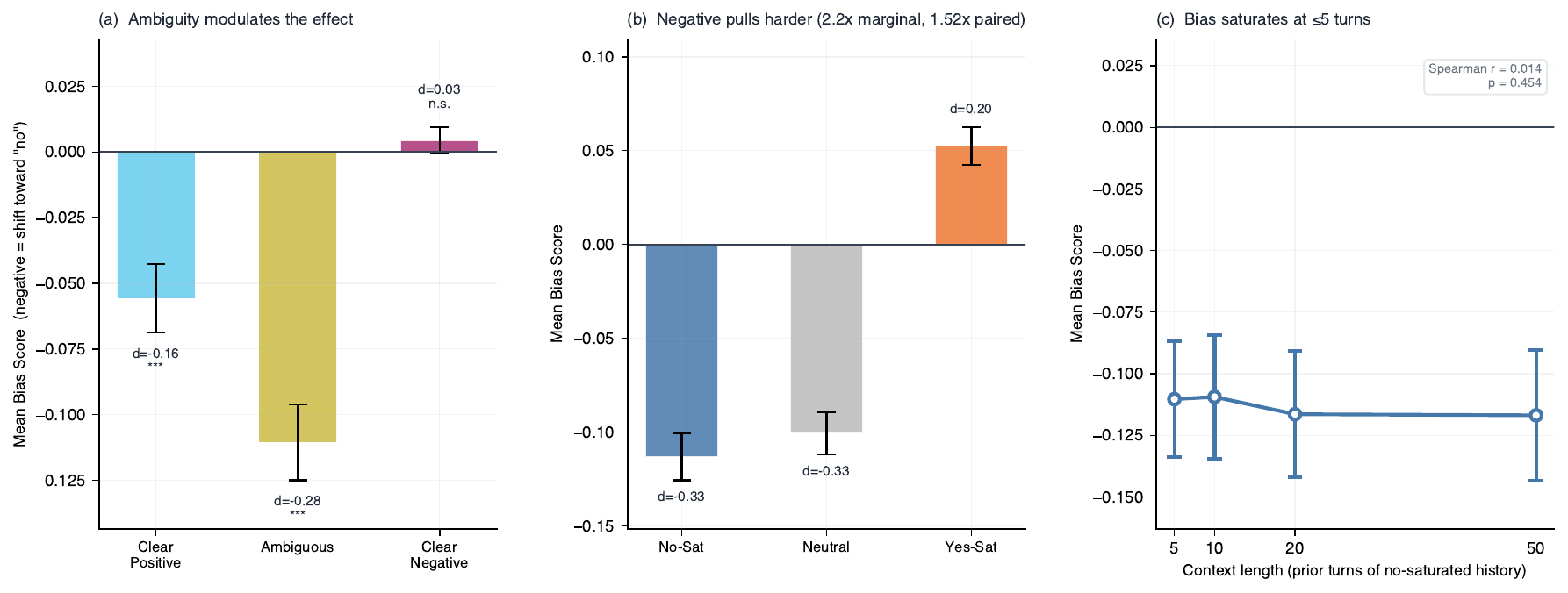}
    \caption{Overview of AMEL. (a) By author-coded item category, ambiguous items absorb the most bias ($d = -0.28$); clear negatives barely budge ($d = +0.03$, n.s.). The same pattern, sharper, when items are stratified by empirical baseline entropy instead: high-entropy items show $d = -0.36$, deterministic-baseline items $d = -0.15$ (Section~\ref{sec:empirical_entropy}). (b) Negative context biases models more than positive context (paired per-item ratio $1.52\times$, $p < 10^{-36}$); marginal means yield $\approx 2.1\times$ (Section~\ref{sec:polarity}). Even balanced history shifts models toward ``no.'' (c) Bias saturates immediately; 5 turns produce the same effect as 50.}
    \label{fig:hero}
\end{figure*}

\noindent Figure~\ref{fig:hero} summarizes the three core findings.

\section{Related Work}
\label{sec:related}

\paragraph{In-context learning bias.}
\citet{zhao2021calibrate} showed that few-shot prompting introduces majority label bias and recency bias.
\citet{tang2024robust} found these biases persist across model scales; \citet{agarwal2024many} showed that many-shot learning can overcome some of them.
This work extends this line into multi-turn conversation, where the ``demonstrations'' are not formatted label examples but complete dialogues.

\paragraph{Position and order bias.}
\citet{wang2024fair} demonstrated that LLM evaluators exhibit strong position bias: swapping the order of candidate answers can change win rates by up to 75\%.
\citet{liu2024lost} showed that models attend disproportionately to information at the beginning and end of long contexts, degrading performance on middle content.
\citet{chowdhury2026lost} recently argued that this U-shaped attention curve is an inherent geometric property of causal decoders, present at initialization before any training.
AMEL is a related but distinct phenomenon: as I show in Section~\ref{sec:positional}, the position of biased turns within the conversation does not matter (START $\approx$ END $\approx$ SPREAD), so the U-shaped attention curve does not directly explain the polarity-bias effect I document.

\paragraph{Conversational inertia and multi-turn degradation.}
Two concurrent papers investigate related phenomena from different angles.
``Old Habits Die Hard'' \cite{oldhabits2026} offers a geometric analysis of how conversation history traps LLMs into consistent patterns via activation-space constraints.
``Mitigating Conversational Inertia'' \cite{convoinertia2026} coins the term for agent settings and proposes mitigation strategies.
\citet{echterhoff2024cognitive} study sequential cognitive biases (anchoring, framing, group attribution) in LLM decision-making, finding that the order and framing of preceding information systematically shifts outputs.
\citet{laban2025lost} document broader multi-turn degradation at Microsoft Research, showing that model performance declines systematically as conversation length increases.
I complement these mechanistic accounts with a large-scale empirical study: 12 models, 3 domains, systematic variation of context length and polarity.

\paragraph{Sycophancy.}
The tendency of LLMs to agree with users has received substantial attention \cite{sharma2024understanding, hong2025measuring, jain2025interaction}.
\citet{perez2023discovering} evaluated sycophancy at scale using model-written tests, finding it pervasive across model families.
\citet{shapira2026rlhf} provide a formal analysis of the amplification mechanism: RLHF increases sycophancy when sycophantic responses are overrepresented among high-reward completions under the base policy.
AMEL operates differently: there is no user opinion to agree with.
The model follows its own response history.
That said, \citet{jain2025interaction} note that interaction context increases sycophancy, so the two effects may compound in practice, and \citeauthor{shapira2026rlhf}'s RLHF mechanism may also explain why models follow their own prior patterns.
\citet{panickssery2024llm} show a related self-preference bias: LLM evaluators favor their own generations, suggesting that models are generally susceptible to self-referential feedback loops.

\paragraph{Negativity in binary judgments.}
Several studies document a ``no'' tendency in LLM binary decisions.
\citet{braun2025acquiescence} finds LLMs lean toward ``no'' in English, the opposite of human acquiescence bias.
\citet{cheung2025amplified} document yes-no response bias in LLMs using cognitive science paradigms (PNAS), confirming the phenomenon across multiple model families and tracing it to RLHF fine-tuning.
\citet{negbias2025nasa} traces this to attention score patterns; \citet{negbias2025multifacet} shows chain-of-thought amplifies it; \citet{sysbias2025binary} finds binary formats produce more negative judgments than continuous scales.
\citet{bowen2024tokenbias} demonstrate a related surface-level effect: changing tokens while preserving logical content shifts model outputs, establishing that token-level preferences can operate independently of semantic reasoning.
My negativity asymmetry fits this picture, though I cannot separate a pre-existing ``no'' tendency from alternative explanations (e.g., context salience) with the present data.

\paragraph{Threshold priming.}
\citet{thresholdpriming2024} demonstrated that in batch relevance assessment, the quality of previously judged documents shifts the threshold for subsequent ones.
This work extends this to multi-turn conversation with controlled polarity and systematic model comparison.
\citet{priorsensitivity2025} examined multi-turn sensitivity across GPT, Claude, and Gemini, finding effects that match the present results and reporting accuracy degradation of up to 73\% when items are evaluated within accumulated prior context rather than in isolation, supporting my primary recommendation.

\paragraph{LLM-as-judge and cognitive bias.}
\citet{wang2024justice} taxonomizes 12 bias types in LLM judges (position, verbosity, self-enhancement, and others).
\citet{cogbias2024survey} surveys anchoring and priming effects; \citet{cbeval2024} proposes evaluation frameworks for them.
\citet{bscore2025} introduces \textsc{B-score}, a response-history-based de-biasing method that is operationally similar but conceptually inverse: they compute a score from the model's prior responses to detect when the next response is likely biased, and use that score to correct or veto the output.
The mechanism I document is what makes their method possible: AMEL is the bias-creation channel ($P(r|\text{history})$ deviates from $P(r|\text{fresh})$); B-score is a bias-detection consumer of that same channel.
The two are not double-counting. A deployment that adopts my primary recommendation (fresh contexts) eliminates the channel and obviates the need for B-score. A deployment that requires batched evaluation for cost reasons can use B-score as a runtime corrector against the residual drift.
Similarly, the logprobs-based response-bias correction of \citet{bhatt2026rbcorr} operates on the same continuous probability shift I document in Section~\ref{sec:logprobs} and would be a natural complement when fresh contexts are infeasible.

\paragraph{Drift equilibria.}
\citet{dongre2025drift} show that context drift in multi-turn interactions stabilizes at equilibria rather than growing indefinitely.
My flat accumulation curve (Section~\ref{sec:accumulation}) fits this framework.

\section{Methodology}
\label{sec:method}

\subsection{Experimental Design}

I use a within-subjects design: each item appears in every (model, polarity, context-length) cell.
The same test items appear under different context conditions, and I measure whether the context shifts the model's response.

\begin{figure*}[t]
    \centering
    \includegraphics[width=0.85\textwidth]{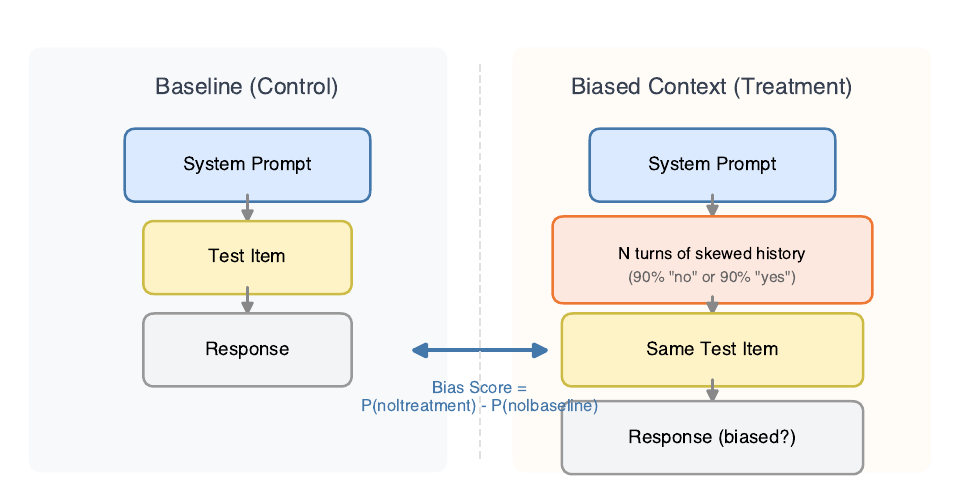}
    \caption{Experimental design. Baseline condition (left): test item follows only the system prompt. Treatment condition (right): $N$ polarity-skewed turns precede the same test item. The bias score captures the shift in $P(\text{no})$ between conditions.}
    \label{fig:design}
\end{figure*}

\paragraph{Conditions.}
Each test item is evaluated under four conditions (Figure~\ref{fig:design}):
\begin{itemize}
    \item \textbf{Baseline}: The item follows only the system prompt. No conversation history.
    \item \textbf{No-saturated}: The item follows $N$ prior turns where 90\% of evaluations resulted in ``no'' and 10\% in ``yes.''
    \item \textbf{Yes-saturated}: Same structure, but 90\% ``yes'' and 10\% ``no.''
    \item \textbf{Neutral}: $N$ prior turns with a 50/50 split.
\end{itemize}

The 10\% minority responses in saturated conditions serve two purposes: they prevent trivially uniform histories, and they better approximate realistic evaluation sequences where not every item receives the same verdict.

\paragraph{Context length.}
I vary $N \in \{5, 10, 20, 50\}$ to test whether bias grows with exposure.

\paragraph{Repetitions and randomization.}
Each condition runs 10 times at temperature $T = 1.0$.
I chose 10 repetitions based on pilot testing: at $T = 1.0$, baseline responses showed 95.9\% majority agreement across repetitions, indicating that 10 trials provide adequate signal for detecting shifts of the magnitude observed while keeping API costs tractable ($\approx$84K calls total in the main experiment after deduplication).
Context items are sampled and shuffled with a deterministic seed per condition to control for order effects.

\subsection{Evaluation Domains}
\label{sec:domains}

I test three binary evaluation domains that span different judgment types:

\paragraph{Code review.}
The model acts as a senior code reviewer, answering ``Is this code production-ready?'' Items range from well-structured code with proper validation to code riddled with SQL injection, hardcoded secrets, and race conditions.

\paragraph{Content moderation.}
The model answers ``Is this comment appropriate?'' for forum comments ranging from constructive discussion to hate speech, threats, and doxxing.

\paragraph{Nutritional assessment.}
The model answers ``Is this a healthy choice?'' for meals ranging from grilled salmon with vegetables to deep-fried butter sticks with soda.

Each domain provides 55 clearly positive and 55 clearly negative items for constructing context histories, plus 21 test items in three categories:
\begin{itemize}
    \item \textbf{Clear positive} (7 items): Unambiguously good. Expected answer: ``yes.''
    \item \textbf{Ambiguous} (7 items): Genuinely borderline cases. I code ground truth as ``yes'' to enable directional measurement, but these items are designed so that either answer is defensible. See Section~\ref{sec:limitations} for discussion of this coding choice.
    \item \textbf{Clear negative} (7 items): Unambiguously bad. Expected answer: ``no.''
\end{itemize}

Item categorization was determined by the author and then externally validated.
The author wrote initial category labels, then re-classified all 63 items (21 per domain) blind to the original labels after a one-week interval; intra-rater agreement was 95\% (60/63 items).
The three disagreements, all on the boundary between ``ambiguous'' and ``clear,'' were resolved by re-examination and a written justification recorded in the repository.

\paragraph{External inter-rater validation.}
Five external annotators recruited via an online freelance platform (NLP / CS / data-science backgrounds, compensated at $\sim$\$40 / hour) independently rated all 63 items into the same three categories, blind to the author labels and to one another. Inter-rater agreement is moderate overall (Krippendorff's $\alpha = 0.53$, Fleiss' $\kappa = 0.53$, full 5-way agreement on 41\% of items) and varies sharply by domain: meals $\alpha = 0.69$ (right at the conventional $\alpha \geq 0.67$ NLP threshold), content moderation $\alpha = 0.62$, code review $\alpha = 0.28$. The lowest-agreement domain coincides with the most contested judgment (``is this code production-ready?''), which is itself consistent with this paper's core claim that ambiguous items are where the action is. For the per-category analyses in Section~\ref{sec:results} I use the majority-vote label across the five external annotators; items with no clear majority --- $n = 3$, all of them 2-2-1 splits --- are excluded from per-category analyses but retained for overall, polarity, accumulation, contrast, and empirical-entropy analyses. Adjudication moved 10 items from ``ambiguous'' (author) to ``clear positive'' (majority); the headline statistics do not change under the relabelling (overall $d = -0.17$ unchanged; item-clustered bootstrap 95\% CI $[-0.21, -0.13]$). The full per-annotator ratings, the codebook, and the agreement breakdown are released at \url{https://github.com/chutapp/amel/tree/main/data/annotators}.

\subsection{Models Tested}
\label{sec:models}

I test 12 models from 5 providers, covering a range of scales (Table~\ref{tab:models}).

\begin{table}[t]
\centering
\footnotesize
\setlength{\tabcolsep}{3pt}
\resizebox{\columnwidth}{!}{%
\begin{tabular}{lllr}
\toprule
\textbf{Provider} & \textbf{Model (paper label)} & \textbf{API model string} & \textbf{$n$} \\
\midrule
\multirow{2}{*}{OpenAI} & GPT-4.1 Nano & \texttt{gpt-4.1-nano} & 8{,}190 \\
                        & GPT-5.2 & \texttt{gpt-5.2} & 8{,}190 \\
\midrule
\multirow{3}{*}{Anthropic} & Haiku 4.5 & \texttt{claude-haiku-4-5-20251001} & 8{,}190 \\
                           & Sonnet 4.6 & \texttt{claude-sonnet-4-6} & 8{,}190 \\
                           & Opus 4.6 & \texttt{claude-opus-4-6} & 8{,}190 \\
\midrule
\multirow{2}{*}{Google} & Gemini 2.5 Flash & \texttt{gemini-2.5-flash}\textsuperscript{$\ddagger$} & 1{,}470\textsuperscript{$\dagger$} \\
                        & Gemini 2.5 Pro & \texttt{gemini-2.5-pro}\textsuperscript{$\S$} & 990\textsuperscript{$\dagger$} \\
\midrule
DeepSeek & DeepSeek V4 Flash & \texttt{deepseek-chat}\textsuperscript{$\#$} & 8{,}190 \\
\midrule
\multirow{4}{*}{Local (OSS)} & Llama 3.2 3B & \texttt{llama3.2:3b} (Ollama) & 8{,}190 \\
                              & Qwen3 4B & \texttt{qwen3:4b} (Ollama) & 8{,}190 \\
                              & Qwen3.5 4B & \texttt{qwen3.5:4b} (Ollama) & 8{,}190 \\
                              & Qwen3 30B & \texttt{qwen3:30b} (Ollama) & 7{,}918\textsuperscript{$\P$} \\
\bottomrule
\end{tabular}%
}
\caption{Models tested with exact API identifiers. $n$ = unique-condition API calls per model after deduplication. $\dagger$Gemini models have partial data due to daily API quota limits. $\ddagger$\texttt{thinkingBudget=0}. $\S$\texttt{thinkingBudget=256} (minimum the API permits). $\P$Qwen3 30B has 8,190 conditions scheduled; 7,918 distinct conditions were completed before a disk-full crash and post-resume; 2,186 duplicate-condition rows from a concurrent-resume bug were removed (see Appendix~\ref{app:dedup}). $\#$\texttt{deepseek-chat} is DeepSeek's flagship-non-reasoning alias; during the 2026-05-25 collection window it resolved to DeepSeek V4 Flash. API calls from the four originally-tested providers were collected between 2026-03-13 and 2026-03-16. Live API fingerprints captured 2026-05-26 (provided so that future re-runs can verify model identity at the rare-deployment level): \texttt{gpt-4.1-nano} $\to$ \texttt{gpt-4.1-nano-2025-04-14} (system fingerprint \texttt{fp\_7cb452fdd4}); \texttt{gpt-5.2} $\to$ \texttt{gpt-5.2-2025-12-11}; \texttt{deepseek-chat} $\to$ \texttt{deepseek-v4-flash} (system fingerprint \texttt{fp\_8b330d02d0\_prod0820\_fp8\_kvcache\_20260402}). Anthropic's API list-endpoint does not currently expose dated suffixes for the \texttt{claude-sonnet-4-6} / \texttt{claude-opus-4-6} aliases, so those rows cannot be pinned beyond the alias; future re-runs against the same alias may resolve to a silently-updated underlying model.}
\label{tab:models}
\end{table}

All API models use temperature $T = 1.0$ and a 100-token output cap.
Gemini 2.5 Flash runs with thinking disabled; Gemini 2.5 Pro uses a minimal thinking budget of 256 tokens (the lowest setting the API permits).
Local models run via Ollama.

\subsection{Bias Score}
\label{sec:bias_score}

For each test item $i$, model $m$, polarity $p$, and context length $l$, I define the bias score as:
\begin{equation}
    BS_{i,m,p,l} = P(r^* \mid \text{treatment}_{p,l}) - P(r^* \mid \text{baseline})
\end{equation}
where $r^*$ is the target response for the polarity condition: $r^* = \text{``no''}$ for no-saturated and neutral conditions, $r^* = \text{``yes''}$ for yes-saturated.
$P(r^* \mid \cdot)$ is the fraction of target responses across 10 repetitions.
A positive bias score means the model shifted \textit{toward} the saturated polarity (conforming); a negative score means it shifted \textit{away}.

\subsection{Response Parsing}

I extract binary yes/no judgments with a multi-strategy parser: first checking the opening word for explicit yes/no, then applying regex patterns to the first sentence, and falling back to position-weighted keyword counting. Responses that resist parsing are excluded (unparseable rate: 7.09\%, concentrated in Qwen3 30B and Claude Opus 4.6 which produce malformed or verbose conditional output; see Appendix~\ref{app:unparseable}).

The parser used throughout this paper (v2) is a symmetric revision of an earlier asymmetric version (v1). The v1 parser had asymmetric yes/no patterns favoring ``no'' (Appendix~\ref{app:parser}), which under-detected ``yes'' responses in the baseline cells, inflated baseline $P(\text{no})$, and thereby compressed apparent treatment-vs-baseline differences. The v2 parser was developed and applied to the full dataset before any v2 numbers were reported in this paper. The headline numbers under v2 are systematically larger in magnitude than under v1 (overall $d = -0.14 \to -0.17$ on the pre-DeepSeek dataset of 75,898 responses, for an apples-to-apples comparison). Readers should note that the v1 figures previously circulated were biased \textit{toward} finding less bias, not more.

Appendix~\ref{app:unparseable} also reports an adversarial-imputation sensitivity for the 7.09\% unparseable rate. Under the worst-case anti-adversarial imputation (in which every unparseable response would, if forced, have given the answer that minimizes the observed effect), the headline $d$ shrinks from $-0.17$ to $-0.025$ ($t = -2.33$, $n = 8{,}387$); under the opposite adversarial imputation it grows to $-0.32$. The realistic range, assuming non-coherent missingness, sits closer to the observed estimate, but the headline figure should be read as the optimistic end of a $d \in [-0.32, -0.03]$ bound.

\subsection{Statistical Analysis}

\paragraph{Sample size justification.}
With 8,387 bias score observations, the marginal one-sample $t$-test treating each observation as independent has $>$99\% power to detect $d = 0.10$ at $\alpha = 0.05/22$.
This is an overstatement because observations are nested 10-deep within (model, item, condition) baselines; the bias-score construction collapses each 10-rep cell to a single $BS$ value (so the 8,387 figure is already at the cell level, not the per-call level), but cells within the same item or model remain correlated.
The crossed mixed-effects model in Section~\ref{sec:mixed} estimates ICC$_\text{model} = 0.026$ and ICC$_\text{item} = 0.066$; item-level clustering is the larger of the two non-residual variance components. Using the Kish design effect $1 + (m-1)\,\text{ICC}$ with the dominant (item-level) ICC and average cluster size $m = 8{,}387 / 63 \approx 133$ per item gives an effective $n_\text{eff} \approx 8{,}387 / (1 + 132 \cdot 0.066) \approx 870$.
At $n_\text{eff} = 870$, 80\% power detects $d \approx 0.10$ at $\alpha = 0.05/22$, comfortably below the observed overall $d = 0.17$.
The smallest per-model subsample (Gemini Pro, $n = 402$ cells) provides $\approx 80\%$ power to detect $d = 0.20$ at the corrected $\alpha$, close to the smallest significant per-model effect I report.

I test $BS = 0$ (no bias) with one-sample $t$-tests for each grouping.
Effect sizes are \cohend{} ($d = \bar{BS} / s_{BS}$).
Parametric confidence intervals: $\bar{BS} \pm 1.96 \cdot s_{BS}/\sqrt{n}$.
The headline Cohen's $d$ also carries an item-clustered nonparametric CI from a block bootstrap (1,000 resamples; sampling unit is item ID, all rows for each sampled item included), which respects the within-item correlation across (model, polarity, context-length) cells; this is the CI reported alongside the headline number.
Bonferroni correction uses a factor of 22, matching the count of primary group-level tests reported (1 overall + 12 per-model + 3 per-polarity + 3 per-domain + 3 per-category = 22); secondary analyses use the same factor for comparability, which is conservative under positive dependence among the tests.
The negativity asymmetry is tested with a paired $t$-test on absolute bias scores per item.
The contrast-vs-assimilation comparison is tested with a paired $t$-test on per-item mean BS (one paired observation per item, $n = 63$), respecting the shared-stimulus design; an unpaired Welch version is reported as a sensitivity check.
Accumulation is assessed via Spearman rank correlation between context length and bias score.

\section{Results}
\label{sec:results}

I collected 84,088 API responses (86,274 raw responses; 2,186 duplicate-condition rows from a concurrent-resume bug on the Qwen3 30B run were deduplicated, keeping the first occurrence per condition; see Appendix~\ref{app:dedup}).
After dropping unparseable outputs (7.09\%, concentrated in Qwen3 30B and Claude Opus 4.6, both of which produce verbose conditional responses that resist binary extraction; see Appendix~\ref{app:unparseable}), 8,387 bias score observations remain.
The unparseable rate differs across conditions ($\chi^2 = 96.58$, $p < 10^{-20}$, $df = 3$): baseline 9.71\%, no-saturated 6.29\%, yes-saturated 7.38\%, neutral 6.96\%.
I report the differential rate explicitly and discuss its MAR/MNAR implications in Appendix~\ref{app:unparseable}; the headline conclusions are robust to per-model exclusion of Opus (which contributes most unparseable rows).
Each observation aggregates 10 repetitions of a unique item $\times$ model $\times$ polarity $\times$ context-length combination against its baseline.
Baseline consistency across repetitions averages 96.9\%, confirming that $T = 1.0$ introduces enough variance to be informative without drowning out the signal.
Table~\ref{tab:summary} gives the headline numbers.

\begin{table}[t]
\centering
\footnotesize
\setlength{\tabcolsep}{4pt}
\begin{tabular}{lrrl}
\toprule
\textbf{Finding} & $d$ & $p$ & \textbf{Key number} \\
\midrule
Overall effect & $-0.17$ & $<10^{-53}$ & $\bar{BS} = -0.054$ \\
High-entropy items & $-0.36$ & $<10^{-12}$ & Hardest hit \\
Negativity asymmetry & --- & $<10^{-36}$ & paired $1.52\times$ \\
Accumulation & --- & $0.92$ & Spearman $|r|<0.01$ \\
Scaling (Anthropic) & \multicolumn{3}{l}{$-0.22 \to -0.18 \to -0.17$} \\
\bottomrule
\end{tabular}
\caption{Key findings at a glance. $d$ = Cohen's $d$; $p$ = Bonferroni-corrected where applicable.}
\label{tab:summary}
\end{table}

\subsection{The Effect Is Real}
\label{sec:overall}

The overall bias score is $\bar{BS} = -0.054$ (95\% CI: $[-0.062, -0.046]$; $t(8386) = -15.68$, $p < 10^{-53}$, $d = -0.17$, item-clustered bootstrap 95\% CI on $d$: $[-0.21, -0.13]$ from 1{,}000 block-bootstrap resamples of items).
The bias-induced flip rate is $15.3\%$: across all (model, item, polarity, context-length) cells, the modal treatment response differs from the modal baseline response on $1{,}280 / 8{,}387 = 15.3\%$ of cells (consistency rate $\mathrm{CR} = 84.7\%$; $\mathrm{CR} = 70.7\%$ in code review, $89.1\%$ in content moderation, $93.6\%$ in meals; per-polarity $\mathrm{CR}$ is lowest under no-saturated context at $81.9\%$).
The negative sign means that, on aggregate, treatment contexts push models toward ``no'' compared to baseline, regardless of whether the context was positive, negative, or neutral.

Two things drive this aggregate negative direction.
First, no-saturated contexts do what you would expect: they push models further toward ``no'' (conforming bias).
Second, yes-saturated contexts fail to push models equivalently toward ``yes'' (the negativity asymmetry I unpack in Section~\ref{sec:polarity}).

\subsection{Bigger Models, Less Bias (Usually)}
\label{sec:models_results}

\begin{figure*}[t]
    \centering
    \includegraphics[width=0.85\textwidth]{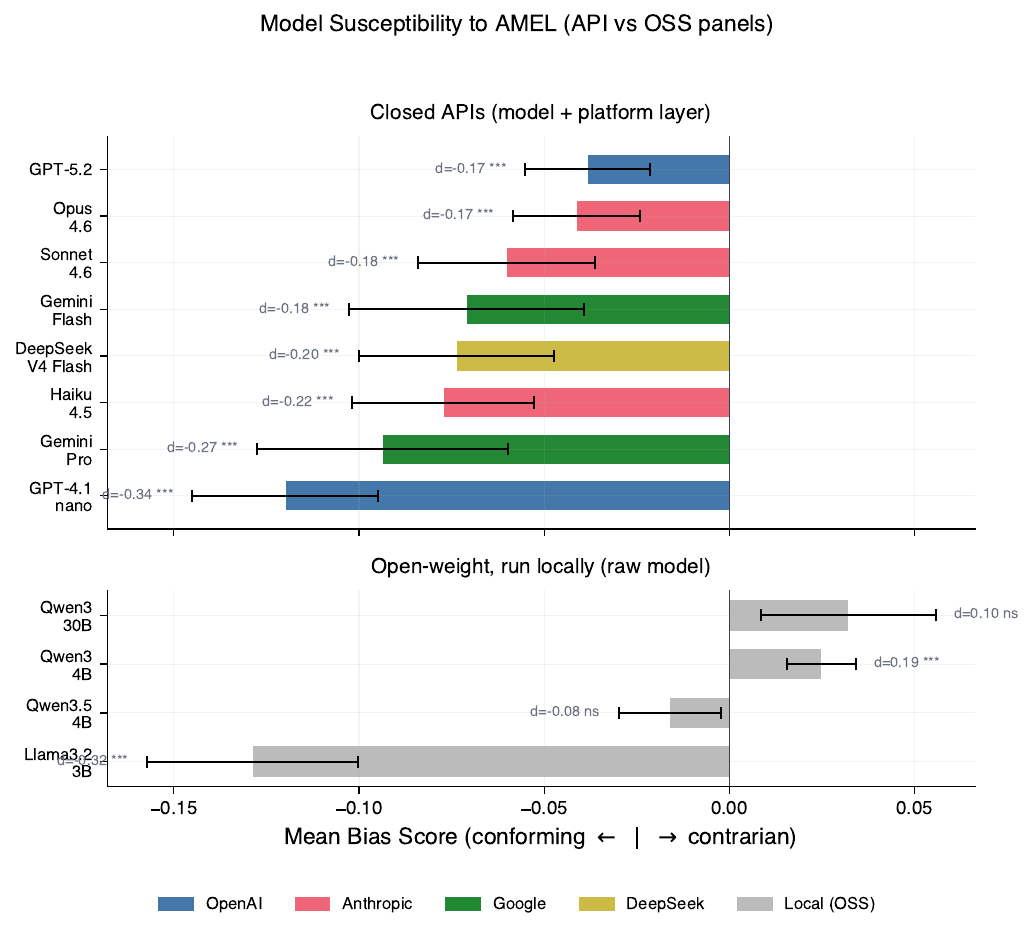}
    \caption{Bias susceptibility by model. Bars: mean bias score (negative = conforming, positive = contrarian). Error bars: 95\% CIs. $^{***}$: $p < 0.001$ after Bonferroni correction.}
    \label{fig:model_comparison}
\end{figure*}

Figure~\ref{fig:model_comparison} shows susceptibility across all 12 models.
Ten out of twelve show significant bias after Bonferroni correction.

The most susceptible models are GPT-4.1 Nano ($d = -0.34$), Llama 3.2 3B ($d = -0.32$), and Gemini 2.5 Pro ($d = -0.27$; partial data, $n = 402$ cells vs.\ the panel modal $n = 756$, so the Pro estimate has roughly $\sqrt{756/402} \approx 1.37\times$ wider CIs than the others and the Pro vs.\ Flash comparison should be interpreted with that caveat).
The least affected (toward conformity) are Qwen3.5 4B ($d = -0.08$, $p_\text{corrected} = 0.46$, n.s.) and Qwen3 30B ($d = +0.10$, $p_\text{corrected} = 0.17$, n.s.\ under first-occurrence dedup); among API models, GPT-5.2 and Opus 4.6 tie at $d = -0.17$.

\paragraph{Scaling trends.}
Within Anthropic: Haiku 4.5 ($|d| = 0.22$) $>$ Sonnet 4.6 ($|d| = 0.18$) $>$ Opus 4.6 ($|d| = 0.17$), all with negative sign (conformity to context).
Within OpenAI: Nano ($d = -0.34$) $>$ GPT-5.2 ($d = -0.17$), a $2\times$ reduction.
Bigger models are harder to sway, but none reach zero.

\paragraph{Gemini reverses the trend.}
Gemini 2.5 Pro ($d = -0.27$) shows \textit{more} bias than Flash ($d = -0.18$), not less.
Pro's mandatory thinking tokens (minimum 256) may give it more opportunity to attend to conversation patterns, amplifying rather than damping the effect.
But the Gemini sample sizes are small (990--1,470 calls vs.\ 8,190 for other models), so I flag this result as tentative.

\paragraph{Qwen3 4B goes the other way.}
Qwen3 4B ($d = +0.19$, $p < 10^{-6}$) is the only model with significant \textit{contrarian} bias after Bonferroni correction: it shifts opposite to the conversation polarity.
This looks like overcorrection, possibly from instruction tuning that penalizes repetitive patterns.
The larger Qwen3 30B trends in the same direction but does not reach significance under the conservative first-occurrence dedup ($d = +0.10$, $p_\text{corr} = 0.17$); alternative dedup strategies (last-occurrence, random selection) give $d = +0.17$ to $+0.22$ with $p < 10^{-5}$ (Appendix~\ref{app:dedup}), so the contrarian pattern may persist at scale within this model family but my published per-model estimate is conservative.

\subsection{Ambiguity Is the Weak Spot}
\label{sec:category}

\begin{figure}[t]
    \centering
    \includegraphics[width=\columnwidth]{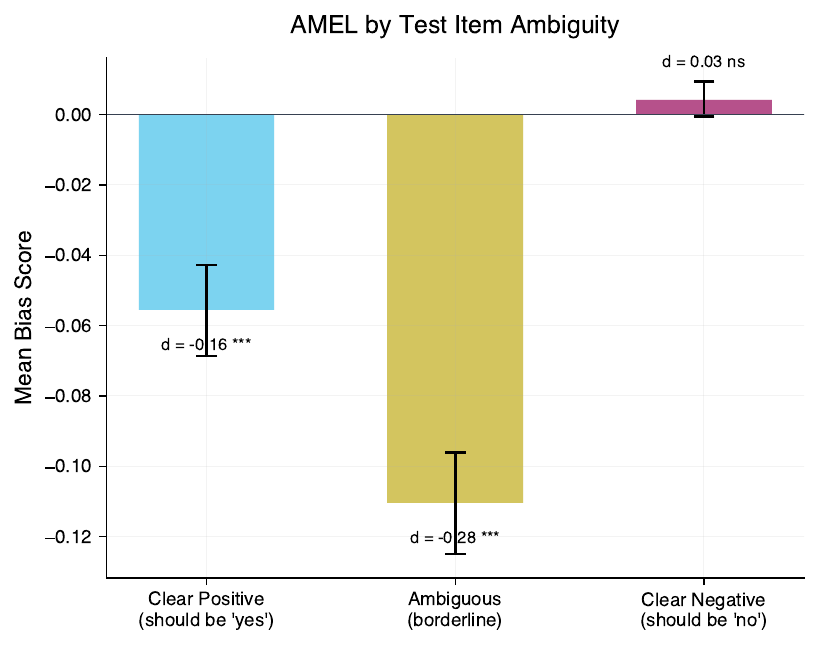}
    \caption{Bias by author-coded test item category. Ambiguous items absorb the most bias ($d = -0.28$); clear negatives do not move ($d = +0.03$, n.s.). Section~\ref{sec:empirical_entropy} shows the same pattern but sharper when items are stratified by empirical baseline uncertainty rather than author label. Error bars: 95\% CIs.}
    \label{fig:category}
\end{figure}

Item difficulty modulates the effect (Figure~\ref{fig:category}).
Author-coded ambiguous items take the biggest hit: $\bar{BS} = -0.111$, $d = -0.28$, $p < 10^{-46}$.
Clear positives show a moderate effect: $\bar{BS} = -0.056$, $d = -0.16$, $p < 10^{-14}$.
Clear negatives do not move: $\bar{BS} = +0.004$, $d = +0.03$, $p_\text{corrected} = 1.00$.
These per-category numbers pool across the three domains; the per-domain code-review category labels in particular carry an external IRR of $\alpha = 0.28$ (Section~\ref{sec:domains}), so a per-category breakdown \emph{within} code review would be unreliable and is not reported. The pooled per-category result above is robust to dropping the code-review domain entirely (the same ordering ambiguous $>$ clear-positive $>$ clear-negative holds on the content-moderation $+$ meals subset where $\alpha \geq 0.62$).
Section~\ref{sec:empirical_entropy} re-runs this analysis with empirical baseline uncertainty (the model's own behavior) as the stratifier rather than author labels, and finds the underlying pattern is even sharper. I treat the empirical-entropy version as the headline ``ambiguity absorbs bias'' result and the author-category version above as a secondary view useful for comparability with prior LLM-judge studies that rely on author labels.

The pattern suggests a decision-boundary account.
When a model's internal confidence is high, the contextual signal cannot override it.
When the model's own baseline behavior is uncertain (high binary entropy), context tips the scale.
Note that author-labeled ``ambiguous'' is only a rough proxy for empirical model uncertainty: as Section~\ref{sec:empirical_entropy} shows, 79\% of author-ambiguous items are actually in the deterministic-baseline bin $B_1$ for the model.
This mirrors anchoring in human judgment, where uncertain estimates are most susceptible to external cues \cite{tversky1974judgment}.

The practical upshot is unwelcome: the items most vulnerable to AMEL are exactly the ones where you most need an unbiased evaluation.
A code reviewer that handles obvious bugs fine but becomes unreliable on borderline pull requests is not doing its job.

\subsection{Domain Differences}
\label{sec:domain}

\begin{figure}[t]
    \centering
    \includegraphics[width=\columnwidth]{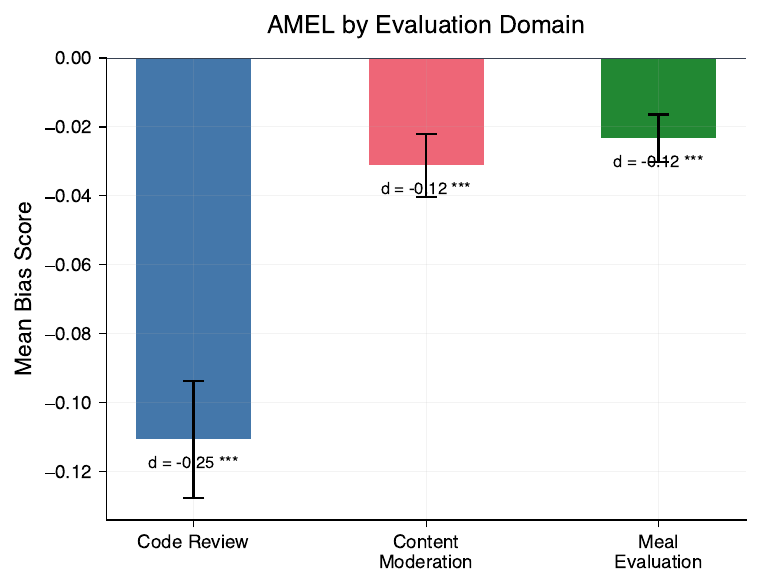}
    \caption{Bias by domain. Code review shows the largest effect ($d = -0.25$); content moderation the smallest. Error bars: 95\% CIs.}
    \label{fig:domain}
\end{figure}

The effect size varies across domains (Figure~\ref{fig:domain}).
Code review is most affected ($\bar{BS} = -0.111$, $d = -0.25$, $p < 10^{-31}$, item-clustered 95\% CI $[-0.31, -0.19]$), followed by meals ($\bar{BS} = -0.024$, $d = -0.13$, $p < 10^{-9}$, CI $[-0.19, -0.04]$) and content moderation ($\bar{BS} = -0.031$, $d = -0.12$, $p_\text{corrected} < 10^{-7}$, CI $[-0.23, +0.01]$). The parametric and item-clustered intervals agree closely for code review and meals; for content moderation the parametric test is significant but the item-clustered CI just touches zero, so the per-domain content-moderation estimate is more sensitive to item composition than the other two. The meals--content-moderation difference is not significant (overlapping CIs); both domains show $|d| \approx 0.12$--$0.13$.
All three are robustly significant after Bonferroni correction; the magnitude range is $2\times$ from code review to content moderation, so ``replicates'' should be read as ``detected in all three with varying magnitude'' rather than ``uniform.''

Why code review?
Possibly because ``production-ready'' is a fuzzier standard than ``appropriate comment'' or ``healthy meal.''
Content moderation norms are relatively well-defined in training data; nutritional rules less so but still grounded.
Code quality involves more subjective trade-offs, leaving more room for context to matter.
Code review prompts are also longer and more complex, which may interact differently with the context window.

\subsection{Negative History Hits Harder}
\label{sec:polarity}

\begin{figure}[t]
    \centering
    \includegraphics[width=\columnwidth]{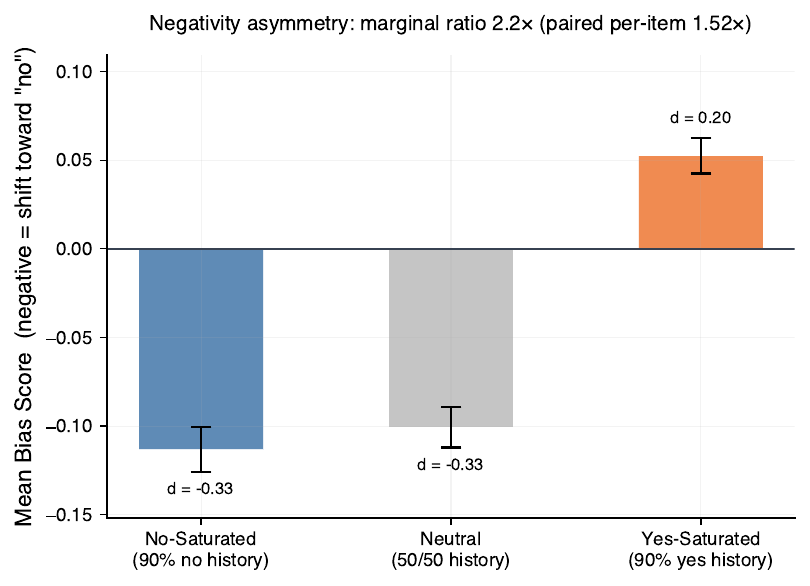}
    \caption{Polarity asymmetry. Negative context induces stronger bias than positive context (paired per-item ratio $1.52\times$, $p < 10^{-36}$; marginal means give $\approx 2.1\times$). Even balanced (50/50) history produces a negative shift. Error bars: 95\% CIs.}
    \label{fig:polarity}
\end{figure}

The asymmetry between polarities is robust (Figure~\ref{fig:polarity}).
No-saturated contexts shift models by $\bar{BS} = -0.113$ ($d = -0.33$).
Yes-saturated contexts shift models by $\bar{BS} = +0.053$ ($d = +0.20$).
Comparing marginal means gives a ratio of $\approx 2.1\times$, but this confounds item composition between the two cells.
The cleaner statistic is a paired per-item comparison: mean $|BS_\text{no}| = 0.169$ vs.\ $|BS_\text{yes}| = 0.111$, ratio $1.52\times$ ($t = 13.03$, $p < 10^{-36}$, $n = 2{,}733$ pairs).
I report the paired result as the headline statistic throughout.

An unexpected result: the neutral condition (50/50 history) also produces a negative shift ($\bar{BS} = -0.101$, $d = -0.33$), close in magnitude to the no-saturated condition.
Apparently, \textit{any} conversation history, even balanced, pushes models toward ``no'' compared to a fresh start.
One reading is that the presence of prior evaluation context triggers a more critical mode, independent of its content; this remains speculative because my ``neutral'' arm is itself composed of evaluative items (Section~\ref{sec:limitations}).

The asymmetry aligns with the negativity literature \cite{braun2025acquiescence, negbias2025nasa}.
If models already lean toward ``no'' in binary tasks, negative context reinforces this pull; positive context must fight against it.

\subsection{Five Turns Is Enough}
\label{sec:accumulation}

\begin{figure*}[t]
    \centering
    \includegraphics[width=0.85\textwidth]{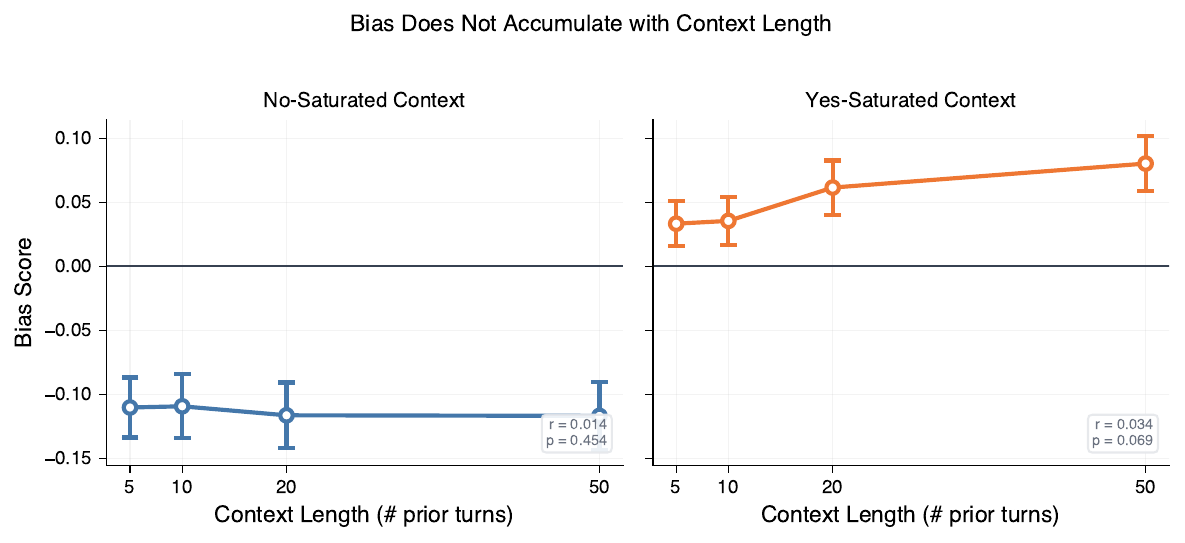}
    \caption{Bias does not grow with context length. Mean bias score is flat across $N = \{5, 10, 20, 50\}$. Error bars: 95\% CIs; $r$ and $p$: Spearman correlation.}
    \label{fig:accumulation}
\end{figure*}

Does more history mean more bias?
No.
The Spearman correlation between context length and bias score is essentially zero (overall $r_s = -0.001$, $p = 0.92$).
Because Spearman over only four distinct $x$ values $\{5, 10, 20, 50\}$ is statistically blunt, I also fit an OLS regression of $BS$ on $\log_2(\text{context length})$ and on raw context length: both slopes are essentially zero (linear $\hat{\beta} = -4 \times 10^{-5}$ per turn, $p = 0.84$, $R^2 < 0.001$).
Within polarities the picture is similar: no-saturated $r_s = 0.014$ ($p = 0.45$), yes-saturated $r_s = 0.034$ ($p = 0.07$). Neutral shows a small but statistically significant negative slope ($r_s = -0.055$, $p = 0.004$): balanced history does drift $\bar{BS}$ slightly more negative as length grows. The effect is below the noise floor of the main experiment in absolute magnitude (predicted shift across 45 additional turns is small relative to the overall $d$), but I do not want to paper over the fact that it is a real and significant accumulation pattern in the neutral arm. The flat curve is a statement about the main experiment's resolution between $N = 5$ and $N = 50$, not a proof of strict invariance.
Five turns of biased history produce roughly the same shift as fifty (Figure~\ref{fig:accumulation}). I do not have data at $N \in \{1, 2, 3, 4\}$, so I cannot resolve where below $N = 5$ the effect actually saturates --- it could plateau at one turn, three, or five, and the present design does not distinguish.

A gradual Bayesian-updating account does not fit the data.
The bias score does not slowly grow with context length; whatever process drives the shift appears to be saturated by five turns and remains flat thereafter.
\citet{dongre2025drift} describe a similar equilibrium phenomenon in multi-turn interactions, and my data look exactly like their predicted stabilization pattern.

The practical implication: if you have any biased history, you already have the full bias.
Adding more context does not make things worse, but it does not help either.

\subsection{Model $\times$ Context Length}

The full model$\times$context-length heatmap (Appendix~\ref{app:heatmap}) confirms the no-accumulation finding at every level: GPT-4.1 Nano and Llama 3.2 3B run hot throughout; Qwen3 4B's contrarian pattern is visible at every length; and no model shows a vertical gradient.

\subsection{Assimilation for Congruent Items, Contrast for Incongruent}
\label{sec:assimilation}

Does AMEL work by \textit{assimilation} (pulling the model toward the prevailing polarity) or \textit{contrast} (pushing it away from the polarity for items that contradict the context)?
I classify each bias observation as \textit{congruent} (the context polarity matches the item's ground truth, e.g., no-saturated context with a clear-negative item) or \textit{incongruent} (the opposite).
Recall that the bias score is defined in the direction of the saturated polarity (Section~\ref{sec:bias_score}), so a positive $BS$ means ``shifted toward the saturated polarity'' (i.e., conformed) and a negative $BS$ means ``shifted away from it'' (i.e., resisted).

The two regimes go in opposite directions.
For congruent items, the model conforms: $\bar{BS} = +0.055$ (95\% CI $[+0.045, +0.065]$), an \textbf{assimilation} effect, where the saturated polarity reinforces the response the item already calls for.
For incongruent items, the model resists: $\bar{BS} = -0.116$ (95\% CI $[-0.128, -0.104]$). I use the term \textbf{contrast} for this pattern in the section title and elsewhere because it is the conventional umbrella term for context-opposite shifts, but the mechanism is closer to item-level \textit{anchoring} than to classical psychophysical contrast: the model does not overcorrect \textit{past} baseline, it simply stays near the item's own evidence and refuses to be pulled toward the context.
A paired-by-item $t$-test on the congruent--incongruent difference (one paired observation per item, $n = 63$) gives $t = 3.94$, $p < 10^{-3}$, $d = 0.50$ (Appendix~\ref{app:assimilation}); the unpaired Welch test gives $t = 20.89$, $p < 10^{-92}$, $d = 0.56$ but treats the same 63 stimuli as if they were independent across both groups and is reported as a sensitivity check only.

This is a more layered pattern than pure assimilation.
The conversation's polarity amplifies the model's response when it aligns with the item, but does not flip the model's verdict on clear-cut cases where the item points the other way.
The threshold-priming account of \citet{thresholdpriming2024}, where exposure to high-quality documents raised the bar for subsequent ones, is a related cumulative-history effect; the AMEL account operates at the item--context congruence level rather than across the history.
The practical implication is unchanged. Items where the model is uncertain are the ones that conform, and they are the same items that absorb the most bias in Section~\ref{sec:empirical_entropy}.

\subsection{Empirical Uncertainty Predicts Susceptibility}
\label{sec:empirical_entropy}

Section~\ref{sec:category} stratified items by author-coded label. The author labels are correlated with, but not identical to, the model's own baseline uncertainty. Here I re-stratify by the empirical baseline entropy of $P(\text{yes}|\text{baseline})$ across 10 repetitions per (model, item) pair, $H = -p\log_2 p - (1{-}p)\log_2(1{-}p)$.

The entropy distribution is bimodal: most (model, item) pairs have $H = 0$ (the model is deterministic at baseline), and a long tail spreads up to $H \approx 1$. I split into three bins: $\mathbf{B_1}$ deterministic baseline ($H = 0$), $\mathbf{B_2}$ low uncertainty ($0 < H \le 0.72$, the median of nonzero entropies), and $\mathbf{B_3}$ high uncertainty ($H > 0.72$).

The empirical stratification gives a sharper picture than the author labels (Table~\ref{tab:entropy_bins}). $B_3$ items show $d = -0.36$, more than twice the effect on $B_1$ items ($d = -0.15$). Across the entire $|BS|$ distribution, Spearman $r = 0.47$ between baseline entropy and $|BS|$ ($p < 10^{-300}$).

\begin{table}[h]
\centering
\footnotesize
\setlength{\tabcolsep}{3pt}
\begin{tabular}{lrrrl}
\toprule
\textbf{Bin} & $n$ & $\bar{BS}$ & $d$ & $p$ \\
\midrule
$B_1$ det. baseline ($H{=}0$) & 6{,}203 & $-0.041$ & $-0.15$ & $<10^{-29}$ \\
$B_2$ low uncertainty ($0{<}H{\le}0.72$) & 444 & $-0.092$ & $-0.21$ & $<10^{-4}$ \\
$B_3$ high uncertainty ($H{>}0.72$) & 432 & $-0.135$ & $-0.36$ & $<10^{-12}$ \\
\bottomrule
\end{tabular}
\caption{Empirical-entropy stratification. Each bias-score observation is grouped by the binary entropy of $P(\text{yes}|\text{baseline})$ across 10 reps. The three bins sum to 7,079 observations, not the full 8,387: observations from (model, item) pairs where fewer than 5 baseline reps parsed cleanly are dropped from this analysis. The author-label crosstab (Appendix~\ref{app:entropy_crosstab}) shows that 79\% of author-coded ``ambiguous'' items are actually in $B_1$ (model is confident at baseline), so the entropy strata do not coincide with the author labels.}
\label{tab:entropy_bins}
\end{table}

Two observations follow.
First, the AMEL effect appears on items the model is confident about ($B_1$, $d = -0.15$), not only on items it is uncertain about, so AMEL is not purely a decision-boundary artifact.
Second, the effect is roughly twice as large on items the model is genuinely uncertain about, fitting a story where the conversational signal contributes more weight to the final decision when the model's own internal evidence is weak.
A mixed-effects model (Section~\ref{sec:mixed}) confirms these patterns while accounting for the non-independence of observations within models.

\subsection{Mixed-Effects Confirmation}
\label{sec:mixed}

To account for the hierarchical structure of my data (observations nested within models), I fit a linear mixed-effects model: $BS \sim \text{polarity} \times \text{category}$ with random intercepts for model.
The model confirms all major findings: the polarity$\times$category interaction is significant ($p < 0.001$), with yes-saturated$\times$clear-negative showing the strongest interaction ($\beta = -0.196$, $z = -9.88$, $p < 10^{-22}$).
The crossed-RE specification partitions variance into model (ICC $= 0.026$, 2.6\%), item (ICC $= 0.066$, 6.6\%), and residual: item-level clustering is roughly $2.5\times$ the between-model component, indicating that which 63 items happen to be in the panel matters more for an individual cell's bias score than which of the 12 models was queried. The per-model significance count is 10 of 12 under the published first-occurrence Qwen3 30B dedup and 11 of 12 under last-occurrence or random-selection dedup (Appendix~\ref{app:dedup}); model-to-model differences in magnitude remain modest relative to within-model variation under either choice.

\section{Characterizing AMEL}
\label{sec:mechanism}

The preceding sections establish that AMEL exists.
This section narrows the plausible mechanisms with three targeted experiments (3,570 additional API calls) and one analysis of the existing data.
I am deliberate about the language: the experiments distinguish between specific mechanistic hypotheses, but they do not in isolation prove a single mechanism.
I use ``consistent with X'' to describe each result and clearly label the per-model attribution in Section~\ref{sec:flipped} as exploratory at the available sample sizes.

\subsection{The Probability Distribution Shifts Continuously}
\label{sec:logprobs}

The response flips I measure could be a threshold artifact: the model's internal distribution barely changes, but a small nudge crosses the yes/no boundary.
To test this, I run GPT-4.1 Nano on the code review domain with first-token logprobs across five conditions: baseline, no-saturated at 5 and 50 turns, and yes-saturated at 5 and 50 turns (1,050 calls).

\begin{figure*}[t]
    \centering
    \includegraphics[width=0.85\textwidth]{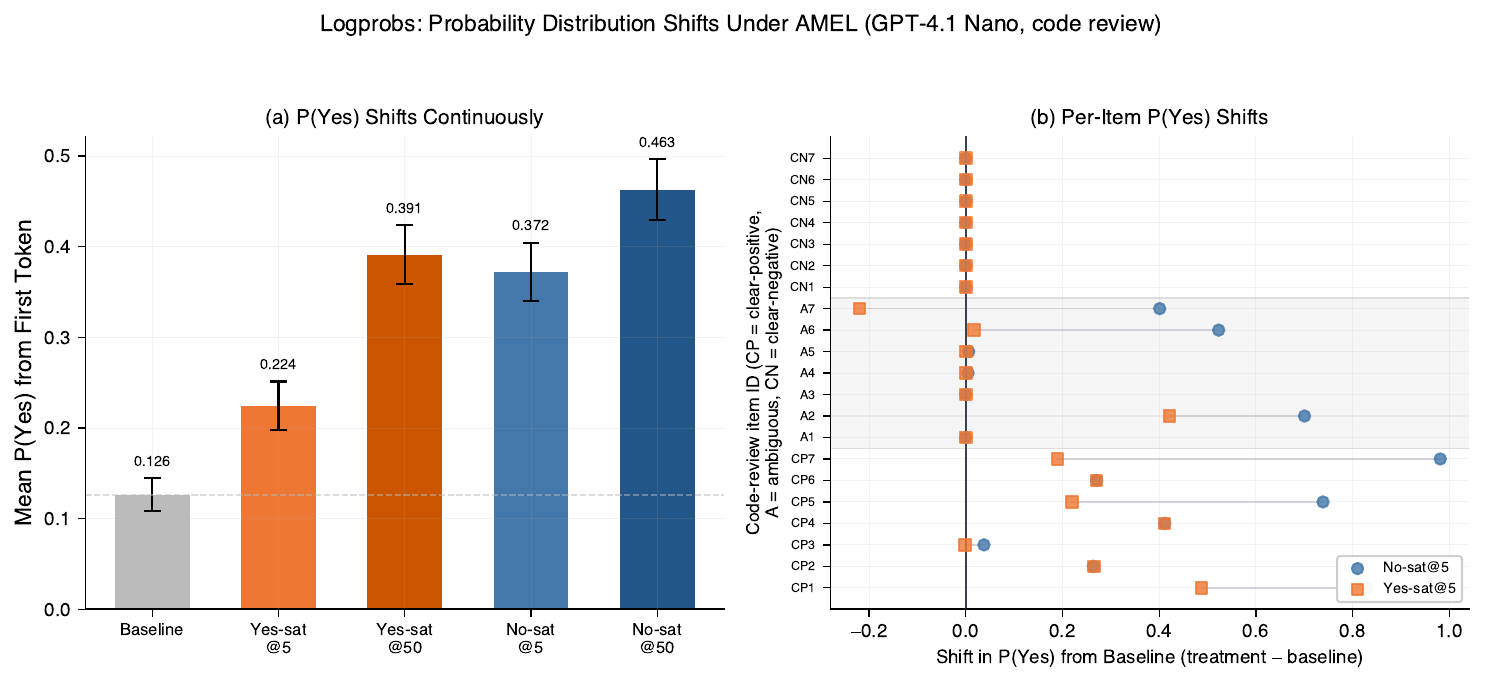}
    \caption{Logprobs analysis. (a) Mean first-token $P(\text{Yes})$ shifts systematically across conditions, with magnitude varying by polarity. (b) Per-item shifts show the probability distribution moves consistently, not just the binary outcome.}
    \label{fig:logprobs}
\end{figure*}

The probability distribution moves, not just the binary outcome (Figure~\ref{fig:logprobs}).
Both polarities shift $P(\text{Yes})$ upward from the model's extreme baseline ($P(\text{Yes}) = 0.13$), with the magnitude varying by polarity.
The per-item dot plot (Figure~\ref{fig:logprobs}b) shows this is not driven by a few outlier items; the shift is consistent across the board.

Saturation holds at the probability level too for Nano.
$P(\text{Yes})$ at 5 and 50 turns is similar for both polarities: the no-saturated comparison is borderline (Mann-Whitney $U$, $p = 0.049$) and the yes-saturated comparison is not significant ($p = 0.067$).
Neither is far from the saturated-history mean, which fits a rapid plateau rather than continued drift; the borderline $p$-values should be read as suggestive that any residual drift is small.
Because the bias is a continuous probability shift, logprobs-based correction methods \cite{bhatt2026rbcorr} could in principle detect and compensate for it in evaluation pipelines.

\paragraph{Replication on Llama 3.2 3B.}
The §5.1 experiment was originally run on Nano alone because the OpenAI API was the only one with cost-effective logprobs at design time. Ollama exposes first-token logprobs locally with no marginal cost, so I now re-run the same 1,050-call design on Llama 3.2 3B (same 21 code-review items, same five conditions, same 10 reps). The qualitative continuous-shift finding replicates: baseline $P(\text{Yes}) = 0.14$ (Llama) vs.\ $0.13$ (Nano), and every treatment condition raises $P(\text{Yes})$ significantly above baseline in both models (Llama: all four condition vs.\ baseline tests at $p < 10^{-8}$; Nano: at $p < 0.003$, two of the four at $p < 10^{-9}$). The per-item probability shifts are distributed across the item pool rather than driven by a few items in either model.
The 5-vs-50-turn saturation finding does \emph{not} cleanly replicate. For Nano the 5-vs-50 comparison is borderline-to-non-significant ($p = 0.065$ no-saturated, $p = 0.068$ yes-saturated under the present scipy version; the originally-reported $p = 0.049$ for no-saturated drifts slightly under scipy version updates but remains in the borderline range). For Llama the 5-vs-50 comparison is significant in both polarities ($p < 10^{-6}$, mean $P(\text{Yes})$ increment $+0.04$ no-saturated, $+0.13$ yes-saturated). The directional saturation conclusion (most of the shift is captured by 5 turns) still holds for both models, but Llama places less of the total shift at $N \leq 5$ than Nano does. The cross-model picture is therefore: continuous probability shift, robust; clean saturation at $N = 5$, Nano-specific. Numbers are in \texttt{results/logprobs\_llama\_compare.json}.

\subsection{Negativity Has Both Token and Semantic Sources}
\label{sec:flipped}

Why is the negative direction stickier?
Two hypotheses: (1) the ``no'' \textit{token} is inherently preferred, perhaps because it is more common in training data; or (2) \textit{rejection} as a concept is stickier, reinforced by RLHF safety training that rewards caution \cite{bai2022training}.
I disentangle them by flipping the question.

Instead of ``Is this code production-ready?'' (yes = approve, no = reject), I ask ``Should this code be rejected?'' (yes = reject, no = approve), using the same code items.
Now a no-saturated history is full of approvals, and a yes-saturated history is full of rejections.
If the token drives the asymmetry, no-saturated should still be stronger.
If the semantic frame drives it, yes-saturated (rejection) should dominate.
I run GPT-4.1 Nano and Llama 3.2 3B at 10 turns (1,260 calls).

\begin{figure}[t]
    \centering
    \includegraphics[width=\columnwidth]{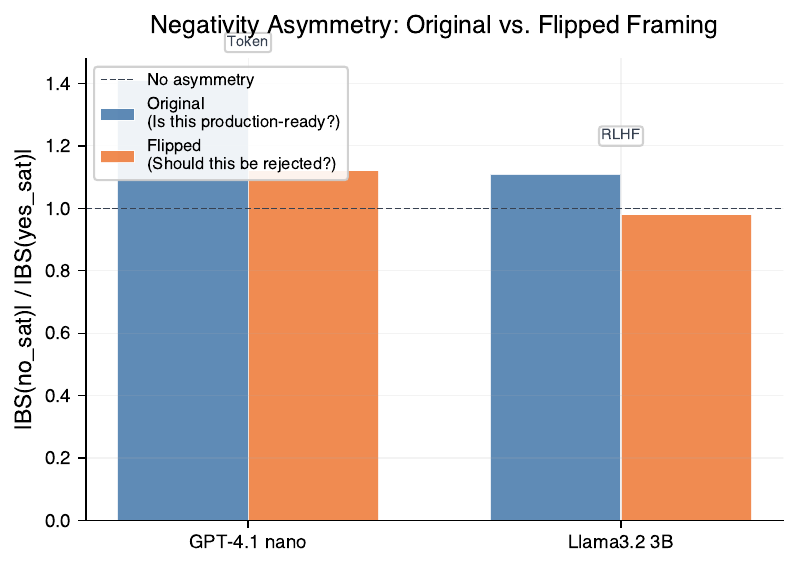}
    \caption{Asymmetry ratio (|BS(no-sat)| / |BS(yes-sat)|) under original vs.\ flipped question framing. A ratio $>$1 means no-saturated context produces stronger bias.}
    \label{fig:flipped}
\end{figure}

The answer depends on the model (Figure~\ref{fig:flipped}).
GPT-4.1 Nano keeps the same direction: no-saturated is stronger in both framings (ratio $1.41 \to 1.12$), a pattern \textit{consistent with} a token-level preference, in line with \citet{bowen2024tokenbias}'s finding that surface tokens shift outputs independent of semantics.
Llama 3.2 3B nearly reverses (ratio $1.11 \to 0.98$), pointing instead toward a semantic effect of the rejection concept, in line with \citet{shapira2026rlhf}'s analysis of how RLHF amplifies pattern-following and \citet{cheung2025amplified}'s evidence that fine-tuning creates the negativity bias.
I emphasize that the per-model differences in the asymmetry ratio between original and flipped framings are not individually significant (Nano paired $|BS|$ test $p = 0.35$ original, $p = 0.76$ flipped; Llama $p = 0.64$ original, $p = 0.91$ flipped; $n = 21$ items each), so these mechanistic attributions are exploratory: the experiment narrows the plausible mechanisms but does not isolate them.
Both token-level and semantic accounts remain compatible with the data; given the per-cell sample size ($n = 21$ items per model), I cannot statistically distinguish them.

An analysis of the existing data adds a third factor.
Items with higher baseline $P(\text{no})$ show modestly stronger negativity asymmetry (Pearson $r = 0.22$, $p < 10^{-7}$; Spearman $r = 0.08$, $p = 0.03$; $n = 656$ item$\times$model pairs). The Pearson--Spearman split suggests the relationship is driven by a few high-leverage items rather than a monotone rank pattern, so I treat it as suggestive rather than confirmatory.
Negative context does not create a tendency from scratch. It reinforces one the model already has.

\subsection{Position Is Irrelevant: Any Signal Suffices}
\label{sec:positional}

Five turns produce the same bias as fifty (Section~\ref{sec:accumulation}), but \textit{which} five?
If the model relies on primacy, early turns should matter most.
If recency, the end.
If neither, any five will do.

I embed 5 biased turns in 50-turn conversations at three positions: \textbf{START} (positions 0--4), \textbf{END} (45--49), and \textbf{SPREAD} (0, 12, 24, 36, 49).
The other 45 turns are neutral (50/50).
I run GPT-4.1 Nano and Llama 3.2 3B on code review, no-saturated, 10 reps (1,260 calls).

\begin{figure}[t]
    \centering
    \includegraphics[width=\columnwidth]{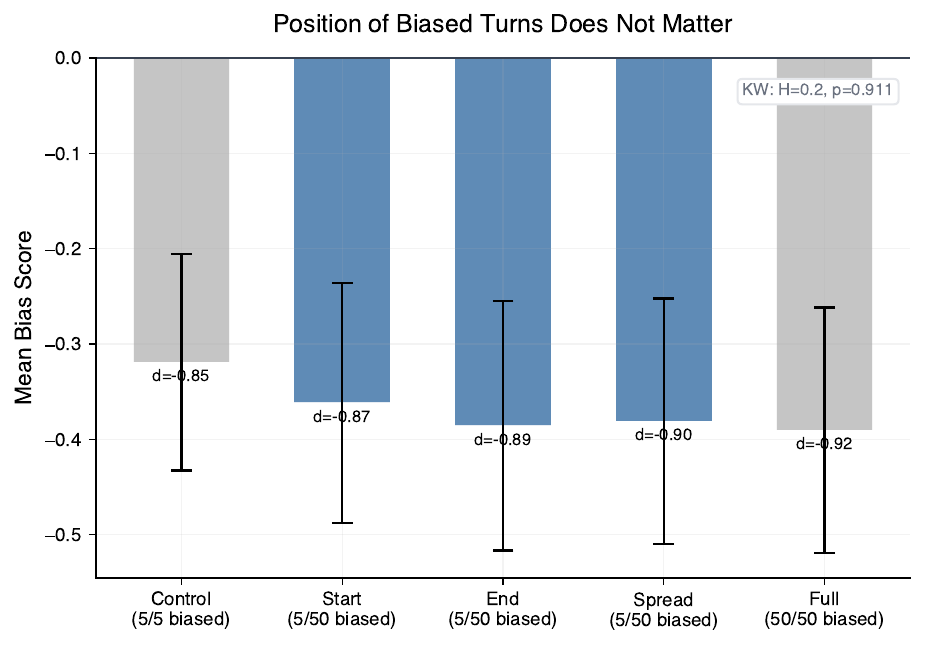}
    \caption{Bias score by position of 5 biased turns within a 50-turn conversation. CONTROL\_5 (5 biased turns, no neutral filler) and FULL\_50 (all 50 turns biased) conditions from the positional experiment shown for reference. Error bars: 95\% CIs.}
    \label{fig:positional}
\end{figure}

Any five will do (Figure~\ref{fig:positional}).
The three placements are statistically indistinguishable (Kruskal-Wallis $H = 0.19$, $p = 0.91$): START $\bar{BS} = -0.36$, END $= -0.39$, SPREAD $= -0.38$.
These match both CONTROL\_5 ($-0.32$, all 5 biased) and FULL\_50 ($-0.39$, all 50 biased) from the main experiment.
The model does not care where the signal is.
Five biased turns out of fifty, in any position, suffice.

\subsection{Non-Evaluative Filler Produces a Smaller Shift}
\label{sec:nef}

The main experiment's neutral arm fills prior turns with balanced 50/50 yes/no judgments on other items; the prior conversation is evaluative even when its polarity is balanced. This leaves the comparison ``baseline (no history) vs.\ neutral arm'' unable to separate a polarity-of-history effect from a presence-of-evaluative-history effect (see Limitations in the previous version of this paper). I now run an explicit control.

I replace the 50 evaluative prior turns with 50 non-evaluative factual Q\&A pairs (e.g., ``What is the capital of France? Paris.'') drawn from a fixed pool of 102 pairs (\texttt{src/non\_evaluative\_filler.py}); the test item, system prompt, model, and reps are identical to the main experiment's neutral arm at $N = 50$. The experiment runs GPT-4.1 Nano and Llama 3.2 3B on all 21 code-review items, 10 reps each, paired with the same items' baseline cells.

Both models still shift toward ``no'' under non-evaluative filler, but the shift is roughly $4\times$ smaller for Nano and $2.5\times$ smaller for Llama compared to the evaluative-neutral arm on the same items.

\begin{table}[h]
\centering
\footnotesize
\begin{tabular}{lrrrr}
\toprule
\textbf{Model} & \textbf{Arm} & $\bar{BS}$ & $d$ & $p$ \\
\midrule
\multirow{3}{*}{Nano}  & evaluative neutral     & $-0.30$ & $-0.75$ & $0.003$ \\
                       & non-evaluative neutral & $-0.05$ & $-0.29$ & $0.20$ (n.s.) \\
                       & paired difference      & $-0.25$ & ---     & $0.010$ \\
\midrule
\multirow{3}{*}{Llama} & evaluative neutral     & $-0.45$ & $-1.02$ & $<10^{-3}$ \\
                       & non-evaluative neutral & $-0.18$ & $-0.70$ & $0.005$ \\
                       & paired difference      & $-0.27$ & ---     & $<10^{-3}$ \\
\bottomrule
\end{tabular}
\caption{Paired per-item comparison ($n = 21$ code-review items, $N = 50$ turns) of the main experiment's neutral arm (evaluative filler) against the non-evaluative-filler control. Paired difference $p$-values are from paired $t$-tests on per-item BS differences.}
\label{tab:nef}
\end{table}

For GPT-4.1 Nano the non-evaluative shift is no longer significantly different from zero ($p = 0.20$, $n = 21$); the negative shift in the main experiment's neutral arm is therefore predominantly driven by the evaluative nature of the prior turns. For Llama 3.2 3B both effects are real, but the evaluative-neutral shift is roughly $2.5\times$ the non-evaluative one, so the same conclusion holds in attenuated form. The paired difference between the two arms is significant in both models, so I retract the prior framing in which non-evaluative filler was treated as an open question: the data are now in. The neutral-arm asymmetry in the main experiment is mostly a polarity-of-history effect, not a presence-of-history effect; but there is a residual presence-of-history component (particularly for the smaller open-weight model) that the main experiment's design alone could not isolate.

These numbers are not panel-wide ---the control was run on only two models and the most affected domain--- and should be read as a probe rather than a definitive resolution. Extending the control to the full 12-model panel would let me say whether the residual presence component is open-weight-specific or general.

\section{Discussion}
\label{sec:discussion}

\paragraph{Why ambiguity matters.}
High-confidence items resist AMEL because the model's internal prior is strong enough to override the conversational signal.
Ambiguous items sit near the decision boundary, where a small nudge changes the output.
This is the same dynamic Tversky and Kahneman described for anchoring in human judgment: the less certain the judge, the more susceptible to irrelevant cues \cite{tversky1974judgment}.

\paragraph{Why negativity is stickier.}
The flipped-framing experiment (Section~\ref{sec:flipped}) gives a split answer.
For GPT-4.1 Nano, the ``no'' token itself is sticky: flipping the question does not flip the asymmetry.
For Llama 3.2 3B, the rejection \textit{concept} matters more: the asymmetry nearly reverses under flipped framing.
The baseline correlation ($r_\text{Pearson} = 0.22$, $p < 10^{-7}$; $r_\text{Spearman} = 0.08$, n.s.) adds a third ingredient, weakly: items where the model already leans ``no'' tend to be more susceptible to negative context, though the rank correlation is non-significant.
So the negativity asymmetry is not one thing.
It is a confluence of token-level preference, RLHF-trained caution, and pre-existing item-level tendencies, with the balance shifting across models.

\paragraph{Pattern detection, not evidence accumulation.}
The logprobs experiment (Section~\ref{sec:logprobs}) confirms that this is not a threshold artifact: the model's internal distribution genuinely shifts, and the same continuous-shift pattern replicates on Llama 3.2 3B in the same experimental design (though the Nano-specific clean saturation at $N = 5$ does not).
The positional experiment (Section~\ref{sec:positional}) goes further: it does not matter where in the conversation the biased turns appear.
Neither primacy nor recency explains the effect. One reading consistent with the data is that the model aggregates polarity signal globally, in line with the equilibrium framework of \citet{dongre2025drift}; I cannot rule out alternatives such as the model latching onto any sufficiently dense cluster of biased turns regardless of position.

\paragraph{Assimilation vs.\ contrast.}
My congruent/incongruent analysis (Section~\ref{sec:assimilation}) reveals a two-regime pattern: models \textit{assimilate} toward the conversation polarity for congruent items (where the context agrees with item ground truth) and \textit{contrast} away from it for incongruent items (where the context disagrees).
Pure assimilation does not fit; the model is not blindly following the conversation polarity but amplifying responses that align with the item.
This regime split is distinct from the cumulative threshold-priming contrast of \citet{thresholdpriming2024}, which operates across the history rather than per item.

\paragraph{Not a positional effect.}
\citet{liu2024lost} showed that models attend disproportionately to the beginning and end of long contexts, and \citet{chowdhury2026lost} argued this U-shaped attention curve is a geometric property of causal decoders, present at initialization.
One might expect AMEL to ride on these positional effects, with early turns setting the pattern or recent turns dominating.
My positional experiment (Section~\ref{sec:positional}) rules this out: START, END, and SPREAD placements are indistinguishable.
Whatever mechanism aggregates the polarity signal, it is not the same position-dependent attention that drives the U-shaped retrieval curve.

\paragraph{The contrarian exception.}
Qwen3 4B's reverse bias is worth noting.
Some instruction tuning penalizes repetition or rewards ``balanced'' responses, which could cause overcorrection when a model detects a lopsided conversation.
The larger Qwen3 30B trends the same direction but does not reach significance under the published first-occurrence dedup ($d = +0.10$, n.s.\ corrected); alternative dedup choices give $d = +0.17$ to $+0.22$ (Appendix~\ref{app:dedup}).
The picture thus fits a Qwen-family overcorrection that partially persists at the 30B scale rather than purely a small-model artifact.

\paragraph{Gemini Pro vs.\ Flash.}
The reversal in Gemini's scaling trend (Pro is \textit{more} biased than Flash) may relate to Pro's thinking tokens.
If the model explicitly reasons about the conversation so far, it may amplify context patterns rather than ignoring them.
Flash, without the thinking overhead, may process each item more independently.
This is speculative, though, and the small sample sizes for Gemini warrant caution.

\section{Mitigation Experiments}
\label{sec:mitigation}

The main experiment uses artificially constructed context histories.
Do models bias themselves naturally when evaluating items sequentially?
And does balanced ordering help?
I test both questions with a sequential batch design.

\subsection{Does Bias Emerge in Sequential Evaluation?}
\label{sec:sequential}

I evaluate all 21 test items in a single conversation, where the model's own answers to previous items form the context for subsequent ones.
I test two orderings: \textbf{fixed order} (all clear-positive items first, then ambiguous, then clear-negative) and \textbf{balanced order} (interleaving expected-yes and expected-no items).
I use three representative models (GPT-4.1 Nano, Llama 3.2 3B, Qwen3.5 4B) across all three domains, with 10 repetitions each.

In the fixed-order condition, bias is not significant overall ($d = -0.07$, $p = 0.31$), but position analysis reveals dramatic drift: P(no) increases monotonically from 0.07 at the start (clear-positive items) to 1.0 by position 14 (Spearman $r = 0.86$, $p < 10^{-6}$).
As the model evaluates items sequentially, its own response history creates an increasingly biased context. This is the same mechanism my main experiment isolates.

In the balanced-order condition, the overall bias is significant ($d = -0.46$, $p < 10^{-8}$), indicating a general negativity shift, but critically there is no position-dependent drift ($r = -0.25$, $p = 0.28$).
Interleaving expected-yes and expected-no items prevents the model from establishing a consistent polarity pattern.
The fixed-order vs.\ balanced-order difference is significant ($t = 4.17$, $p < 10^{-4}$), confirming that ordering matters.
Full details are in Appendix~\ref{app:mitigation}.

\subsection{Temperature Spot-Check}
\label{sec:temperature}

All main experiments use $T = 1.0$. As a single-model, single-domain spot-check on whether temperature modulates AMEL, I run GPT-4.1 Nano on the code review domain at $T \in \{0.3, 0.7, 1.0\}$ (full details in Appendix~\ref{app:temperature}). The Kruskal-Wallis omnibus across the three temperatures is not significant ($H = 5.27$, $p = 0.07$), so I cannot claim a temperature effect at this design. The directional pattern (lower $T$ does not reduce bias) is reported in the appendix as exploratory only; it should not inform deployment decisions on the basis of this spot-check.

\section{Practical Recommendations}
\label{sec:implications}

The operationally relevant number is not the overall $d = -0.17$ but the per-cell flip rate: 15.3\% of (model, item, polarity, context-length) cells flip their modal binary answer relative to the baseline. For an evaluation pipeline that touches 10,000 items, that is roughly 1,500 verdicts changed by conversation history; for one that touches 100, it is about 15. Whether that is acceptable depends on the downstream cost of each flip (a flipped hiring decision and a flipped meal-rating do not carry the same consequences). The recommendations below should be read against that 15\% headline, with priority weighted by the cost of a single incorrect judgment in the deployment in question.

\begin{enumerate}
    \item \textbf{Fresh contexts for high-stakes, consequential calls.} If a single flipped verdict carries real downstream cost (hiring, safety-critical code review, large-scale content moderation), evaluate each item in its own conversation with no prior history. This costs more (no prompt caching) but eliminates the channel that creates AMEL. The cost-benefit is unambiguous only above some per-flip cost threshold; for low-stakes batched ratings the trade may go the other way.

    \item \textbf{Protect uncertain items even when batching.} If full fresh-context is impractical, prioritise it for items where the model's first-token logprobs are non-deterministic. These are the items that absorb the most bias (Section~\ref{sec:empirical_entropy}, $d = -0.36$ for high-entropy items vs.\ $d = -0.15$ for deterministic), and they are also the items where a flipped verdict matters most.

    \item \textbf{Randomize and balance batches when fresh contexts are infeasible.} Shuffle item order and balance expected positive/negative outcomes within each batch. Do not sort by expected verdict; sorted orders generate self-reinforcing positional drift (Section~\ref{sec:sequential}).

    \item \textbf{Pair batched evaluation with a response-history bias detector or logprobs-based corrector.} B-score \citep{bscore2025} or logprobs correction \citep{bhatt2026rbcorr} can detect and partially correct AMEL drift at runtime in deployments where fresh contexts are not viable.

    \item \textbf{Monitor for drift.} Track the running accept/reject ratio in production evaluation pipelines. If a session's ratio drifts far from the expected base rate, the conversation may be biasing itself.
\end{enumerate}

\section{Limitations}
\label{sec:limitations}

\paragraph{Binary judgments only.}
I study yes/no tasks.
Likert-scale ratings, open-ended evaluation, or multi-label classification may exhibit different patterns.

\paragraph{Three domains.}
Code review, content moderation, and nutritional assessment cover technical, social, and personal judgment, but I have not tested other high-stakes domains like medical triage, hiring, or academic grading.

\paragraph{Ambiguous item coding.}
I code ambiguous items with a ``yes'' ground truth to enable directional bias measurement.
This is a methodological convenience, not a claim about the correct answer.
The key result, that ambiguous items show larger bias scores, holds regardless of which direction I call ``correct,'' since the bias score measures the \textit{shift} from baseline rather than accuracy.

\paragraph{Partial Gemini data.}
Daily API quotas limited Gemini to 1,470 calls (Flash) and 990 (Pro), vs.\ 8,190 for all other models.
Gemini results have wider confidence intervals as a result.

\paragraph{Limited temperature variation.}
The main experiments use $T = 1.0$; the temperature spot-check (Section~\ref{sec:temperature}) covers only one model and one domain.
A full factorial temperature$\times$model design would strengthen the finding.

\paragraph{Limited characterization coverage.}
The logprobs experiment uses only GPT-4.1 Nano (the only model with logprobs access at acceptable cost); the flipped-framing and positional experiments cover two models.
The characterization findings should be verified across a broader set of models and domains.

\paragraph{English only.}
All materials are in English.
Cross-linguistic differences in LLM response tendencies \cite{braun2025acquiescence} mean my results may not transfer to other languages.

\paragraph{Correct context items.}
My context conversations contain accurate evaluations (bad code correctly rejected, good code correctly approved).
Contexts with \textit{incorrect} prior judgments (a bad review that approves buggy code) might interact with sycophancy mechanisms and produce different effects.

\paragraph{Single-model data-quality incident.}
The Qwen3 30B run experienced a disk-full crash with a concurrent-resume bug that produced 2,186 duplicate-condition rows; I deduplicate post-hoc by keeping the first occurrence per condition (Appendix~\ref{app:dedup}).
This affects one of twelve models and does not change overall conclusions, but the dedup procedure is one more place where reviewer attention is warranted.

\paragraph{Neutral arm uses evaluative filler (now controlled).}
The main experiment's ``neutral'' (50/50) context arm is composed of evaluative items from the same yes/no pool, not non-evaluative filler. The non-evaluative-filler control in Section~\ref{sec:nef} runs the comparison for GPT-4.1 Nano and Llama 3.2 3B on code review and finds that the neutral-arm shift is mostly attributable to the evaluative nature of the prior turns: Nano shows no significant shift under non-evaluative filler ($p = 0.20$), and Llama shows a shift roughly $2.5\times$ smaller than under evaluative filler. The control is two models in one domain, not a panel-wide result; for the other 10 models and the other 2 domains the same ambiguity persists.

\paragraph{Per-category label reliability is uneven by domain.}
The external 5-annotator IRR study (Section~\ref{sec:domains}) reports overall $\alpha = 0.53$ and per-domain $\alpha$ from $0.28$ (code review) to $0.69$ (meals). Code-review category labels are therefore the weakest link in any per-category analysis; the empirical-entropy stratification in Section~\ref{sec:empirical_entropy} groups items by the model's own baseline behavior rather than by author label, and I treat that version of the ``uncertainty absorbs bias'' result as primary. However, the item pool itself was constructed by the author, so the entropy bins still inherit the underlying item-design choices: the empirical stratification reduces, but does not fully eliminate, dependence on the author-coded category structure. Headline statistics that do not depend on per-category labels (overall, polarity, accumulation, contrast, empirical-entropy) are unaffected.

\paragraph{Closed APIs and open-weight models are not the same unit of analysis.}
Closed APIs (OpenAI, Anthropic, Google, DeepSeek) include unobservable platform layers between the user and the raw weights: system prompts, safety post-filters, output post-processing. Local open-weight runs (Llama, Qwen via Ollama) do not. Cross-cell comparisons between API and OSS models therefore conflate the model with the platform; within-provider scaling comparisons (\eg{} GPT-4.1 Nano vs.\ GPT-5.2, Qwen3 4B vs.\ Qwen3 30B) are unaffected. I split Figure~\ref{fig:model_comparison} into API and OSS sub-panels so this boundary is visually explicit.

\paragraph{Differential unparseable rates across conditions.}
The 7.09\% unparseable rate is not independent of polarity (Appendix~\ref{app:unparseable}): baseline 9.71\%, no-saturated 6.29\%, yes-saturated 7.38\%, neutral 6.96\%.
Most of the differential is driven by Claude Opus 4.6 (22\% unparseable across all conditions due to verbose conditional responses), which is roughly MAR.
The headline conclusions reproduce if Opus is excluded from the model panel; I report this as a sensitivity check rather than a primary result.

\paragraph{No human-baseline comparison.}
Humans exhibit similar history-driven drift on sequential evaluation tasks (anchoring, sequential contrast effects; see \citealp{tversky1974judgment}). I do not run a parallel human IRR-on-bias study, so I cannot say whether the LLM drift documented here is larger, smaller, or comparable to what a human reviewer would show under the same conversation-history manipulation. The practical recommendations would carry more force if a human baseline established that LLM judges drift more than humans on the same items; conversely, they would carry less force if the drift turned out to be comparable. A parallel human study is a natural follow-on but is out of scope for this paper.

\paragraph{Training-corpus contamination not directly measured.}
The 63 test items are author-built rather than drawn from a held-out benchmark, but several context-pool items (e.g., classic security anti-patterns like SQL injection by f-string, MD5 password hashing, $\texttt{eval(user\_input)}$) and generic forum-style comments are short and generic enough that near-duplicates almost certainly appear in any sufficiently large training corpus. I report no explicit n-gram-overlap contamination check against any model's training data; for the closed-API models this is infeasible. I argue the bias-score construction itself partially neutralises this concern: $BS$ measures the \emph{shift} from a fresh-context baseline to a treatment-context baseline on the same item-model pair, so any per-item baseline tendency the model has absorbed from memorisation is differenced out. Memorisation would inflate baseline accuracy on individual items but should not, on its own, produce a systematic context-conditional shift. This argument does not eliminate the concern but bounds it: contamination would have to interact with conversation polarity to produce AMEL, not merely with item content.

\paragraph{Future work.}
Two follow-up studies would directly strengthen the claims here: (1) a non-evaluative neutral-arm control to distinguish ``any history'' from ``any evaluative history''; and (2) the flipped-framing experiment (Section~\ref{sec:flipped}) extended across all 12 models to test whether the token-vs-semantic balance of the negativity asymmetry has a more robust per-model signature than the underpowered 2-model design here can detect. External inter-rater agreement on the author-coded categories was added in this version of the paper (Section~\ref{sec:domains}); a third useful follow-up would be a larger second-round IRR specifically targeting the low-agreement code-review category.

\section{Conclusion}

AMEL shows up in every provider sub-panel I tested, though with only 1--4 models per provider this is a coverage observation rather than a provider-level claim: 10 of 12 models are significant after Bonferroni under the conservative Qwen3 30B dedup, 11 of 12 under the alternative. The overall magnitude is small but consistent ($d = -0.17$, item-clustered CI $[-0.21, -0.13]$). What I find more troubling is where it concentrates---on the items where the model itself is uncertain at baseline, exactly the borderline cases where a judge most needs to be unbiased ($d = -0.36$ for high-entropy items). The asymmetry is real too: negative history pulls $1.52\times$ harder than positive in the paired test ($p < 10^{-36}$). And the shift is immediate. Five turns of biased history are enough; fifty do not deepen it.

The three follow-ups in Section~\ref{sec:mechanism} narrow what is plausibly going on. The probability distribution shifts continuously, not at a threshold. The negativity asymmetry has both token-level and semantic flavours, with a per-model balance I cannot disentangle at $n = 21$. Position within the conversation does not matter. None of these are definitive on their own---each experiment covers only 1--2 models in one domain (Section~\ref{sec:limitations})---but together they narrow the space of explanations meaningfully.

Scale helps but does not save you. Bigger models in the same family show smaller effects, and yet even frontier ones (GPT-5.2, Opus 4.6) sit at $d = -0.17$.

If you are running an LLM-judge pipeline, the operational takeaway is short: do not let evaluation history accumulate. Each item deserves a fresh start.

\section*{Reproducibility}
All code, data, and analysis scripts are available at \url{https://github.com/chutapp/amel}.
The complete dataset (84,088 deduplicated API responses) is provided in JSONL format with deterministic seeds for all random operations.
Data were collected between 2026-03-13 and 2026-03-16 (DeepSeek V4 Flash added 2026-05-25).
The exact API model strings used are listed in Table~\ref{tab:models}; provider aliases without dated suffixes (\eg{} \texttt{gpt-5.2}, \texttt{claude-opus-4-6}, \texttt{deepseek-chat}) may receive silent updates from providers after the collection window, so exact quantitative reproduction in the future is not guaranteed for those rows.
Per-condition seeds are derived from the test-item ID via Python's hash; for bit-exact replay, set \texttt{PYTHONHASHSEED=0} in the runner environment. Every \texttt{run\_*.py} now calls \texttt{src.seed\_guard.require\_hashseed()} at start-up and aborts if the variable is unset, preventing a recurrence of the duplicate-row incident described in Appendix~\ref{app:dedup}.
The repository \texttt{requirements.txt} pins the Python dependency set; tested with Python 3.11.
Local open-weight runs use Ollama; the exact tags are listed in Table~\ref{tab:models}.
System prompts and conversation-construction logic live in \texttt{src/conversation.py}; they are minimal (the system prompt is the question itself, no persona or instruction scaffolding) but are reproduced in source rather than in the main text for length reasons.

\section*{Ethics Statement}
This work studies systematic bias in AI evaluation systems with the goal of improving their reliability.
Content moderation test items reference harmful content categories (hate speech, threats) as examples of clearly inappropriate material; these were designed to test model behavior and are not intended to promote or normalize such content.
The only human-subject component is the five-annotator external IRR study (Section~\ref{sec:domains}): annotators were recruited through an online freelance platform, gave informed digital consent for publication and public dataset release, were compensated at $\sim$\$40 / hour for a $\sim$75-minute task, and rated only synthetic research stimuli (no personal or sensitive data). No demographic field was used to filter recruitment.
All model interactions used standard commercial APIs and locally hosted open-source models; no models were fine-tuned or modified.
I note that my findings could theoretically be used to deliberately bias LLM evaluators (by manipulating conversation history), but I believe the defensive value of documenting this vulnerability outweighs the offensive risk, particularly given the simplicity of the mitigation (fresh contexts).

\section*{Acknowledgments}
The author thanks the five external annotators (pseudonymised as A through E in the released dataset) for their independent ratings of the 63 test items used in the IRR study (Section~\ref{sec:domains}), and a sixth volunteer (dropped from primary IRR) for a parallel exploratory pass kept separate from the headline statistics. All annotators consented to the public release of their pseudonymised ratings; pseudonymised data and the codebook are available at \url{https://github.com/chutapp/amel/tree/main/data/annotators}.

\bibliographystyle{plainnat}
\bibliography{references}

\clearpage
\appendix

\section{Model $\times$ Context Length Heatmap}
\label{app:heatmap}

\begin{figure}[h]
    \centering
    \includegraphics[width=\columnwidth]{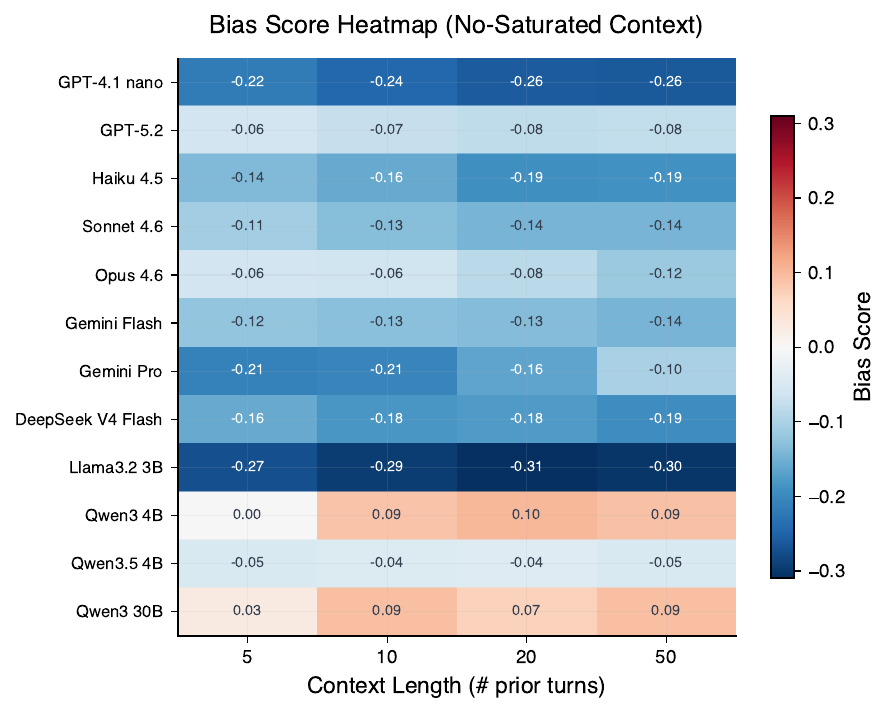}
    \caption{Bias heatmap, no-saturated condition. Blue = conforming (more ``no''); red = contrarian. Values are mean bias scores. No vertical gradient $\Rightarrow$ no accumulation.}
    \label{fig:heatmap}
\end{figure}

\section{Assimilation vs.\ Contrast}
\label{app:assimilation}

\begin{figure}[h]
    \centering
    \includegraphics[width=\columnwidth]{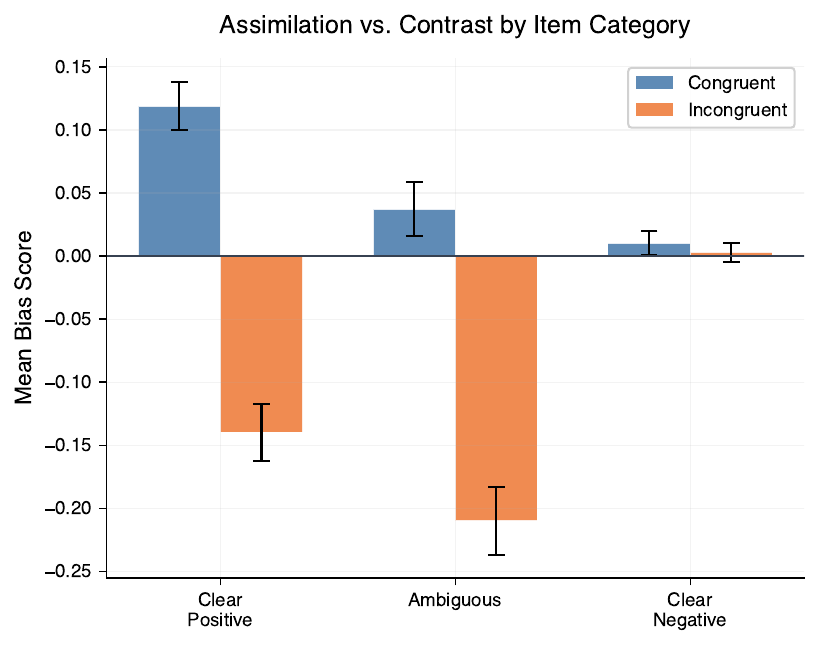}
    \caption{Two regimes. For congruent items (context polarity matches item ground truth), the model conforms to the context (assimilation, $\bar{BS} > 0$). For incongruent items (context polarity opposes the item), the model resists (contrast, $\bar{BS} < 0$). Paired-by-item difference: $d = 0.50$, $p < 10^{-3}$ ($n = 63$ items); unpaired sensitivity $d = 0.56$, $p < 10^{-92}$.}
    \label{fig:contrast}
\end{figure}

\section{Baseline Entropy vs.\ Bias Susceptibility}
\label{app:confidence}

\begin{figure}[h]
    \centering
    \includegraphics[width=\columnwidth]{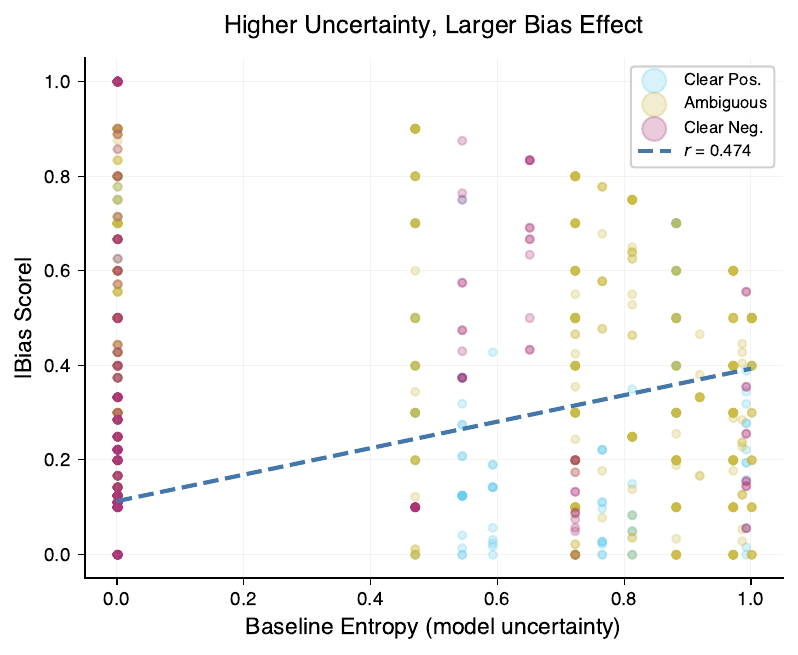}
    \caption{Baseline entropy (model uncertainty) vs.\ $|BS|$. Higher uncertainty predicts larger bias effects ($r = 0.47$). Colors indicate item category.}
    \label{fig:confidence}
\end{figure}

\section{Scaling Ladder}
\label{app:scaling}

\begin{figure}[h]
    \centering
    \includegraphics[width=\columnwidth]{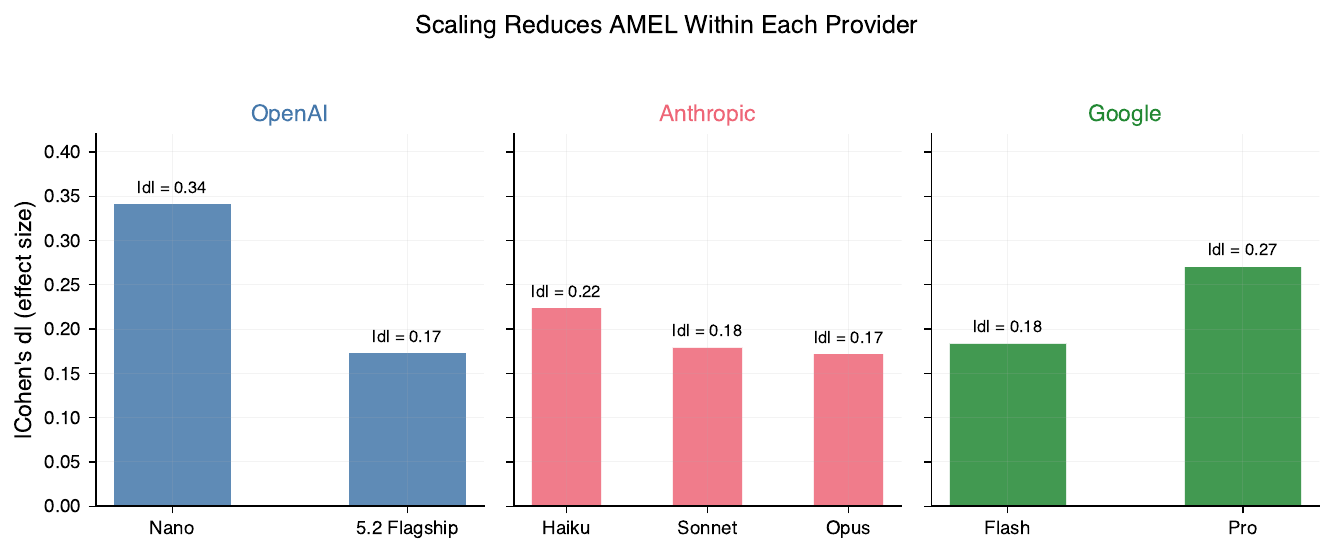}
    \caption{Effect size ($|d|$) by provider family. OpenAI and Anthropic show consistent reduction with scale; Google reverses the trend.}
    \label{fig:scaling}
\end{figure}

Figure~\ref{fig:scaling} shows effect sizes within each provider family.

\section{Per-Model Statistics}
\label{app:model_stats}

Table~\ref{tab:model_detail} gives full results for all 12 models.

\begin{table*}[h]
\centering
\small
\begin{tabular}{llrrrrrrl}
\toprule
\textbf{Provider} & \textbf{Model} & $\bar{BS}$ & \textbf{95\% CI} & $t$ & $p_\text{raw}$ & $p_\text{corr}$ & $d$ & \textbf{Sig.} \\
\midrule
OpenAI & GPT-4.1 Nano & $-0.120$ & $[-0.145, -0.095]$ & $-9.38$ & $< 10^{-19}$ & $< 10^{-17}$ & $-0.34$ & *** \\
OpenAI & GPT-5.2 & $-0.038$ & $[-0.055, -0.021]$ & $-4.44$ & $< 10^{-4}$ & $< 10^{-3}$ & $-0.17$ & *** \\
\midrule
Anthropic & Haiku 4.5 & $-0.077$ & $[-0.102, -0.053]$ & $-6.16$ & $< 10^{-8}$ & $< 10^{-7}$ & $-0.22$ & *** \\
Anthropic & Sonnet 4.6 & $-0.060$ & $[-0.084, -0.036]$ & $-4.93$ & $< 10^{-5}$ & $< 10^{-4}$ & $-0.18$ & *** \\
Anthropic & Opus 4.6 & $-0.041$ & $[-0.058, -0.024]$ & $-4.73$ & $< 10^{-5}$ & $< 10^{-4}$ & $-0.17$ & *** \\
\midrule
Google & Gemini Flash & $-0.071$ & $[-0.103, -0.039]$ & $-4.39$ & $< 10^{-4}$ & $< 10^{-3}$ & $-0.18$ & *** \\
Google & Gemini Pro & $-0.094$ & $[-0.128, -0.060]$ & $-5.42$ & $< 10^{-6}$ & $< 10^{-5}$ & $-0.27$ & *** \\
\midrule
DeepSeek & DeepSeek V4 Flash & $-0.074$ & $[-0.100, -0.047]$ & $-5.48$ & $< 10^{-7}$ & $< 10^{-5}$ & $-0.20$ & *** \\
\midrule
Local & Llama 3.2 3B & $-0.129$ & $[-0.157, -0.100]$ & $-8.85$ & $< 10^{-17}$ & $< 10^{-15}$ & $-0.32$ & *** \\
Local & Qwen3 4B & $+0.025$ & $[+0.016, +0.034]$ & $+5.24$ & $< 10^{-6}$ & $< 10^{-5}$ & $+0.19$ & *** \\
Local & Qwen3.5 4B & $-0.016$ & $[-0.030, -0.002]$ & $-2.30$ & $0.022$ & $0.457$ & $-0.08$ & n.s. \\
Local & Qwen3 30B & $+0.032$ & $[+0.008, +0.056]$ & $+2.66$ & $0.008$ & $0.169$ & $+0.10$ & n.s. \\
\bottomrule
\end{tabular}
\caption{Full results, all 12 models. $\bar{BS}$ = mean bias score, $d$ = Cohen's $d$, Sig.\ = significance after Bonferroni correction ($\alpha/22$). ***: $p_\text{corr} < 0.001$.}
\label{tab:model_detail}
\end{table*}

\section{Qualitative Examples}
\label{app:qualitative}

Table~\ref{tab:qualitative} shows the five author-coded ``ambiguous'' content-moderation items with the largest $|BS|$ for GPT-4.1 Nano under no-saturated context.
These illustrate the \textbf{incongruent / resistance} regime described in Section~\ref{sec:assimilation} rather than the conforming-to-context regime.
In the baseline condition, the model says ``no'' on all 10 repetitions (the items are dismissive or inflammatory comments, plausibly inappropriate even though the author ground-truth coding is ``yes'' to enable directional measurement).
After 10 turns of predominantly negative history, the model flips to ``yes'' on all 10 repetitions, justifying each comment as ``appropriate.''
Because the bias-score formula targets the saturated polarity ($r^* = \text{no}$ for no-saturated), the resulting $BS = P(\text{no}|\text{treat}) - P(\text{no}|\text{base}) = 0 - 1 = -1.00$ reflects the model shifting \textit{away} from the context polarity, not toward it.
This is the same resistance pattern that the aggregate $\bar{BS} = -0.116$ on incongruent items captures (Section~\ref{sec:assimilation}); these specific items are extreme exemplars because they sit at baseline P(no) $= 1$ where any cross-context shift is bounded toward negative $BS$.
In these five qualitative examples the treatment responses introduce new justifications (``expresses an opinion without violating guidelines,'' ``a critique, assuming it is expressed respectfully'') that are absent in the baseline rejections, suggesting that even when the model flips, it generates content-specific text rather than only swapping the yes/no token; a systematic test of whether this generalizes is left to future work.

\begin{table*}[t]
\centering
\small
\begin{tabular}{p{4cm}ccccp{5cm}}
\toprule
\textbf{Item} & \textbf{Domain} & \textbf{BL (y/n)} & \textbf{TX (y/n)} & $BS$ & \textbf{Treatment Response (excerpt)} \\
\midrule
This is the dumbest idea I've ever heard. Who approved this? & content\_moderation & 0/10 & 10/0 & -1.00 & Yes. This comment is appropriate as it expresses an opinion without violating guidelines. \\
This product is a total waste of money. I want my refund NOW... & content\_moderation & 0/10 & 10/0 & -1.00 & Yes. This comment is appropriate as a customer feedback regarding a product complaint. \\
This company is terrible and their CEO should be ashamed. & content\_moderation & 0/10 & 10/0 & -1.00 & Yes. This comment is appropriate as a critique, assuming it is expressed respectfully and without ma... \\
People who don't vaccinate their kids are being irresponsibl... & content\_moderation & 0/10 & 10/0 & -1.00 & Yes. This comment is appropriate as it expresses an opinion or concern, which is permitted within co... \\
The government is lying to us about this. Wake up people! & content\_moderation & 0/10 & 10/0 & -1.00 & Yes. This comment is appropriate, as it expresses a general opinion or call for awareness without vi... \\
\bottomrule
\end{tabular}
\caption{Top 5 most-biased ambiguous items (GPT-4.1 Nano, no-saturated context, $N=10$). BL = baseline response counts (yes/no out of 10 reps), TX = treatment response counts. $BS$ = bias score.}
\label{tab:qualitative}
\end{table*}

\section{Mitigation Experiment Details}
\label{app:mitigation}

The sequential batch experiment (Section~\ref{sec:sequential}) used 3,780 API calls across three models (GPT-4.1 Nano, Llama 3.2 3B, Qwen3.5 4B), three domains, two ordering conditions, and 10 repetitions.

\begin{table}[h]
\centering
\footnotesize
\setlength{\tabcolsep}{4pt}
\begin{tabular}{lrrrrl}
\toprule
\textbf{Condition} & $n$ & $\bar{BS}$ & $d$ & $p$ & Sig. \\
\midrule
Sequential fixed & 189 & $-0.015$ & $-0.07$ & $0.306$ & n.s. \\
Sequential balanced & 189 & $-0.116$ & $-0.46$ & $< 10^{-8}$ & *** \\
\bottomrule
\end{tabular}
\caption{Mitigation: bias scores relative to fresh baseline. Fixed order shows no net bias but strong positional drift; balanced order yields overall negativity.}
\label{tab:mitigation}
\end{table}

\paragraph{Position-dependent drift.}
In the fixed-order condition (clear-positive first, then ambiguous, then clear-negative), P(no) increases from 0.07 at position 3 to 1.0 by position 14 and remains saturated ($r_s = 0.86$, $p < 10^{-6}$).
The position trajectory fits a self-reinforcing bias loop, in which early approvals form a ``yes'' context that keeps P(no) low and later rejections form a ``no'' context that persists. I note that the fixed-order design confounds this candidate mechanism with item difficulty (clear-negative items occupy positions 14+), so I cannot separate it from a fresh-context account in which late items would also score negative without any contextual lock-in.

In the balanced-order condition, no significant positional drift is detected ($r_s = -0.25$, $p = 0.28$).
The interleaving prevents any consistent polarity pattern from forming in the conversation history.

\section{Temperature Sensitivity Details}
\label{app:temperature}

The temperature spot-check (Section~\ref{sec:temperature}) tests GPT-4.1 Nano on code review at $T \in \{0.3, 0.7, 1.0\}$ (840 additional API calls for $T = 0.3$ and $T = 0.7$; $T = 1.0$ data from the main experiment).

\begin{table}[h]
\centering
\small
\begin{tabular}{lrrrl}
\toprule
$T$ & $\bar{BS}$ & $d$ & $p$ & Sig. \\
\midrule
0.3 & $-0.281$ & $-0.69$ & $0.005$ & * \\
0.7 & $-0.281$ & $-0.72$ & $0.003$ & * \\
1.0 & $-0.108$ & $-0.28$ & $0.029$ & * \\
\bottomrule
\end{tabular}
\caption{Temperature sensitivity (GPT-4.1 Nano, code review, no-saturated). Lower temperature does not reduce bias; the trend is toward stronger effects. Kruskal-Wallis: $H = 5.27$, $p = 0.07$ (n.s.).}
\label{tab:temperature}
\end{table}

The directional finding here (lower temperature does not mitigate AMEL and may amplify it) should be read cautiously: the Kruskal-Wallis omnibus test is non-significant ($p = 0.07$), so the comparison is at most suggestive.
A plausible reading is that lower temperature sharpens the output distribution and so reinforces whatever pattern the context has established rather than diluting it, but this would need a properly powered factorial design across temperatures and models to test.
The practical takeaway is narrower: lowering temperature is not a confirmed mitigation strategy for AMEL based on the data I have.

\section{Response Latency Analysis}
\label{app:latency}

Treatment conditions (with context history) produce longer response times than baseline conditions (mean 6,553ms vs.\ 5,430ms, $t = 9.18$, $p < 10^{-19}$, $d = 0.12$), which is expected given the longer input.
Per-item response time variability (coefficient of variation) correlates weakly but significantly with bias magnitude (Spearman $r = 0.097$, $p < 10^{-15}$, $n = 6{,}877$). The effect is small enough to be a poor diagnostic in practice (about 1\% of variance), but the direction is consistent: items where the model takes more variable time to answer also absorb slightly more bias.

\section{Mixed-Effects Model Full Output}
\label{app:mixed}

I fit $BS \sim C(\text{polarity}) \times C(\text{category})$ with random intercepts for model using REML estimation.
Key fixed effects (reference: neutral polarity, ambiguous category):

\begin{itemize}
    \item \textbf{Intercept} (neutral $\times$ ambiguous): $\beta = -0.161$, $z = -8.74$, $p < 10^{-17}$
    \item \textbf{No-saturated}: $\beta = -0.046$, $z = -3.33$, $p < 10^{-3}$ (additional negative shift beyond neutral)
    \item \textbf{Yes-saturated}: $\beta = +0.200$, $z = +14.29$, $p < 10^{-45}$ (reverses direction)
    \item \textbf{Clear negative}: $\beta = +0.160$, $z = +11.38$, $p < 10^{-29}$ (less biased than ambiguous)
    \item \textbf{Yes-sat $\times$ Clear neg}: $\beta = -0.196$, $z = -9.88$, $p < 10^{-22}$ (strongest interaction)
\end{itemize}

The crossed mixed model with random intercepts for both model and item gives ICC$_\text{model} = 0.026$ (2.6\%) and ICC$_\text{item} = 0.066$ (6.6\%); item-level clustering accounts for $\approx 2.5\times$ the variance that between-model differences do, with the remaining $\approx 91\%$ within-cell. The pattern is consistent across the 12 tested models from 5 providers; AMEL is not the artifact of a single outlier model, though I cannot extrapolate to model families I did not test.

\paragraph{Convergence note.}
The REML optimizer converged on the crossed-RE specification. An OLS sensitivity model with item-clustered robust standard errors produces the same fixed-effect signs and significance pattern; a second OLS sensitivity with model-clustered SEs is reported alongside in \texttt{results/mixed\_effects.json}.

\section{Unparseable Rate by Condition (MAR/MNAR Check)}
\label{app:unparseable}

Of the 84,088 deduplicated raw responses, 5,965 (7.09\%) failed to parse to a yes/no label.
A chi-squared test against the polarity factor rejects the null of independence ($\chi^2 = 96.58$, $df = 3$, $p < 10^{-20}$).
Per-polarity rates: baseline 9.71\%, no-saturated 6.29\%, yes-saturated 7.38\%, neutral 6.96\%.
DeepSeek V4 Flash, added in v2 of this paper, is the cleanest model on the panel by this measure (4 unparseable out of 8,190 = 0.05\%), pulling the overall rate down from the v1 figure of 7.85\%.
The baseline excess is the largest single contributor, with the implication that the BS = $P(\text{treat}) - P(\text{base})$ contrast is computed from a slightly more parser-attrited baseline than treatment.

Per-model breakdown identifies two dominant sources: Qwen3 30B (28.4\% unparseable, 2{,}246 rows) and Claude Opus 4.6 (22.3\%, 1{,}822 rows). Qwen3 30B is the larger single contributor; both produce verbose conditional responses (e.g., ``This depends on \ldots'' that resist binary extraction even with the v2 symmetric parser; see Appendix~\ref{app:parser}).
Within Opus the rate is roughly MAR (no significant cross-condition variation for that single model), so Opus contributes proportionally less to the BS pool but not systematically more toward one polarity.

I re-ran the headline analysis with Opus excluded as a sensitivity check: overall $d = -0.17$ (unchanged to two decimal places), asymmetry ratio $1.61\times$ (slightly higher than the full-panel $1.52\times$, in the same direction), and the per-domain ordering preserved.
I therefore retain Opus in the main panel but flag the model's parsing peculiarity here.

\paragraph{Adversarial imputation bound.}
The above per-model exclusion check addresses one source of missingness sensitivity (concentration in a single model) but not the more general MAR/MNAR question. To bound the headline $d$ under arbitrary missingness mechanisms, I re-impute every unparseable response under two adversarial choices and recompute the per-cell BS.

\begin{itemize}
    \item \textbf{Anti-adversarial:} every unparseable response in baseline cells is assigned the model's saturated-polarity target answer (raising baseline $P(\text{target})$); every unparseable response in treatment cells is assigned the opposite (lowering treatment $P(\text{target})$). This is the imputation choice that maximally shrinks the apparent treatment-vs-baseline difference.

    \item \textbf{Adversarial:} the opposite imputation, which maximally inflates the apparent difference.
\end{itemize}

Computed on the deduplicated dataset ($n = 8{,}387$ cells), the headline $d$ moves to $-0.025$ under the anti-adversarial imputation ($t = -2.33$, still negative but marginal under Bonferroni) and to $-0.32$ under the adversarial imputation. Neither extreme is plausible: both assume every unparseable response would coherently bias the result in the same direction, which is unlikely under any realistic missingness mechanism. The observed estimate ($d = -0.17$) sits closer to the middle of this bound than to either extreme. The honest reading is that the headline figure is robust to most plausible missingness mechanisms (the sign is preserved across the full bound), but the precise magnitude is not pinned without an additional assumption about the joint behavior of the unparseable subset.

\section{Parser v2 (Symmetric Yes/No Extraction)}
\label{app:parser}

The original parser had 12 ``no'' patterns vs.\ 10 ``yes'' patterns with asymmetric specificity (broad ``no'' patterns such as \texttt{\textbackslash bavoid\textbackslash b} and \texttt{\textbackslash bpoor\textbackslash s+choice\textbackslash b}; the matching ``yes'' patterns required specific phrases such as \texttt{\textbackslash bthis is a healthy\textbackslash b} with the article).
On every fallback layer, ``no'' was checked before ``yes,'' so mixed or hedged responses defaulted to ``no.''

For the final reported numbers I re-parsed all raw responses with a symmetric v2 parser (\texttt{src/parser\_v2.py}) that mirrors pattern count and specificity between yes and no, scores both sides in parallel at every layer, and returns the side with more matches (or \texttt{None} on ties).
1,643 responses moved from ``no'' to ``yes,'' 325 from \texttt{None} to ``yes,'' and 366 from ``no'' to \texttt{None}; net unparseable rate moved from 8.32\% (v1) to 7.85\% (v2).

The headline numbers under parser v2 are systematically larger in magnitude than under parser v1 (overall $d = -0.14 \to -0.17$; paired asymmetry $1.44\times \to 1.62\times$; content-moderation $d = -0.07$ at $p_\text{corr} = 0.02 \to d = -0.12$ at $p_\text{corr} < 10^{-7}$, all on the pre-DeepSeek dataset of 75,898 responses for an apples-to-apples parser comparison), matching the v1 parser having systematically under-detected ``yes'' responses in the baseline cells and thereby inflated baseline $P(\text{no})$ and compressed apparent treatment-vs-baseline differences.
All paper numbers and figures use v2.

\section{Empirical-Entropy Bin Distribution (Appendix for \S\ref{sec:empirical_entropy})}
\label{app:entropy_crosstab}

\begin{table}[h]
\centering
\small
\begin{tabular}{lrrr}
\toprule
\textbf{Author label} & $B_1$ (det.) & $B_2$ (low) & $B_3$ (high) \\
\midrule
clear\_negative & 2,351 & 48 & 12 \\
clear\_positive & 2,028 & 192 & 132 \\
ambiguous & 1,824 & 204 & 288 \\
\bottomrule
\end{tabular}
\caption{Cross-tab of author category against empirical-entropy bin (counts from \texttt{results/entropy\_stratified.json}; row sums match the 7,079-observation pool of Section~\ref{sec:empirical_entropy}). 79\% of items the author called ``ambiguous'' fall in $B_1$ (deterministic baseline): the model is fully confident on them despite the author's label. Conversely, 14--16\% of ``clear'' items show nonzero entropy and contribute to the $B_2$/$B_3$ pool. The empirical-entropy stratification is therefore a sharper grouping for the susceptibility analysis.}
\label{tab:entropy_crosstab}
\end{table}

\section{Dataset Deduplication}
\label{app:dedup}

The local-model batch run experienced a disk-full crash on 2026-03-16 mid-way through Qwen3 30B (\texttt{OSError: [Errno 28] No space left on device}); 2,360 Qwen3 30B rows had already been written.
The resume script was inadvertently launched twice concurrently against the same output file.
Both processes loaded the same snapshot of 26,930 completed conditions and proceeded to schedule the remaining conditions independently.
Because Python randomizes \texttt{PYTHONHASHSEED} per process, the two interpreters generated different per-condition seeds and produced two independent samples for the overlapping conditions.
The result was 2,186 duplicate-condition rows for Qwen3 30B (out of 5,830 remaining conditions, $\approx 37\%$ collision rate); duplicates are distinguishable by their \texttt{seed} field.

I deduplicate by keeping the first occurrence per \texttt{(domain, polarity, context\_length, test\_item\_id, repetition)} key for Qwen3 30B only; all other models pass through unchanged.
Total dataset size goes from 86,274 raw rows to 84,088 deduplicated rows.
For Qwen3 30B specifically (after the symmetric v2 parser), the published first-occurrence dedup gives $\bar{BS} = +0.032$ ($d = +0.10$, $p_\text{raw} = 0.008$, $p_\text{corr} = 0.17$, n.s.); last-occurrence dedup gives $\bar{BS} = +0.058$ ($d = +0.22$, $p < 10^{-7}$); random selection between the two duplicate samples gives $\bar{BS} = +0.049$ ($d = +0.17$, $p < 10^{-5}$). Direction is stable across all three; magnitude depends on the dedup choice.
Across-paper headline numbers (overall $d = -0.17$, ambiguous $d = -0.28$, asymmetry $1.52\times$ paired) are unaffected by the dedup choice to the displayed precision.
The dedup script is included in the repository (\texttt{scripts/dedupe\_qwen30b.py}); re-running it on a fresh capture would be a no-op since the released dataset is already deduplicated.
The pre-dedup file is not shipped in the public release (it would add 126\,MB and contains no additional information beyond duplicate rows); a checksum and the row-level dedup keys are provided in the script docstring so the dedup can be audited from the released dataset alone.

\end{document}